\documentclass[10pt,twocolumn,letterpaper]{article}

\usepackage{iccv}
\usepackage{times}
\usepackage{epsfig}
\usepackage{graphicx}
\usepackage{amsmath}
\usepackage{amssymb}

\usepackage{subcaption}
\usepackage{algorithm}
\usepackage{algpseudocode}
\usepackage{amsmath}
\usepackage{multirow}
\usepackage{multicol}
\usepackage{url}
\usepackage{color}
\usepackage{bbding}

\usepackage[pagebackref=true,breaklinks=true,letterpaper=true,colorlinks,bookmarks=false]{hyperref}

\newcommand{\tabincell}[2]{\begin{tabular}{@{}#1@{}}#2\end{tabular}}  
\iccvfinalcopy % *** Uncomment this line for the final submission

\def\iccvPaperID{1597} % *** Enter the ICCV Paper ID here

\ificcvfinal\fi

\begin{document}
	
	%%%%%%%%% TITLE
	\title{Weakly-supervised Semantic Segmentation in Cityscape \\ via Hyperspectral Image}
	
	\author{Yuxing Huang\\
		School of Electronic Science and Engineering\\
		Nanjing University\\
		{\tt\small mf1923026@smail.nju.edu.cn}
		\and
		Ying Fu\\
		School of Computer Science and Technology\\
		Beijing Institute of Technology\\
		{\tt\small fuying@bit.edu.cn}

		\and
		Shaodi You\\
		Computer Vision Research Group\\
		University of Amsterdam\\
		{\tt\small s.you@uva.nl}
		\and
		Qiu Shen\\
		School of Electronic Science and Engineering\\
		Nanjing University\\
		{\tt\small shenqiu@nju.edu.cn}
	}
	
	\maketitle
	\ificcvfinal\thispagestyle{empty}\fi
	
	%%%%%%%%% ABSTRACT
	\begin{abstract}
		%Spectral acquisition equipment has developed rapidly in recent years. 
		High-resolution hyperspectral images (HSIs) contain the response of each pixel in different spectral bands, which can be used to effectively distinguish various objects in complex scenes.
		While HSI cameras have become low cost, algorithms based on it have not been well exploited. 
		In this paper, we focus on a novel topic, weakly-supervised semantic segmentation in cityscape via HSIs.
		It is based on the idea that high-resolution HSIs in city scenes contain rich spectral information, which can be easily associated to semantics without manual labeling. Therefore, it enables low cost, highly reliable semantic segmentation in complex scenes.
		Specifically, in this paper, we theoretically analyze the HSIs and introduce a weakly-supervised HSI semantic segmentation framework, which utilizes spectral information to improve the coarse labels to a finer degree.
		The experimental results show that our method can obtain highly competitive labels and even have higher edge fineness than artificial fine labels in some classes. At the same time, the results also show that the refined labels can effectively improve the effect of semantic segmentation. The combination of HSIs and semantic segmentation proves that HSIs have great potential in high-level visual tasks.
	\end{abstract}
	
	%%%%%%%%% BODY TEXT
	\vspace{-12pt}
	\section{Introduction}
	Semantic segmentation in cityscape scenes using RGB images has been well exploited (\emph{e.g}\onedot, FCN~\cite{Long_2015_CVPR}, DeepLabv3~\cite{chen2017rethinking} and HRNet~\cite{wang2019deep}). A various of datasets enable such research. (\emph{e.g}\onedot, Cityscapes \cite{cordts2016the}, CamVid \cite{brostow2009semantic} , BDD100K \cite{yu2018bdd100k} and KITTI \cite{geiger2012we}).
	
\begin{figure}[t]
		\centering
		\begin{minipage}[b]{\linewidth}
			\subfloat[An example of image and metamerism phenomenon.]{
				\begin{minipage}[b]{\linewidth} 
					\centering
					\includegraphics[width=\linewidth]{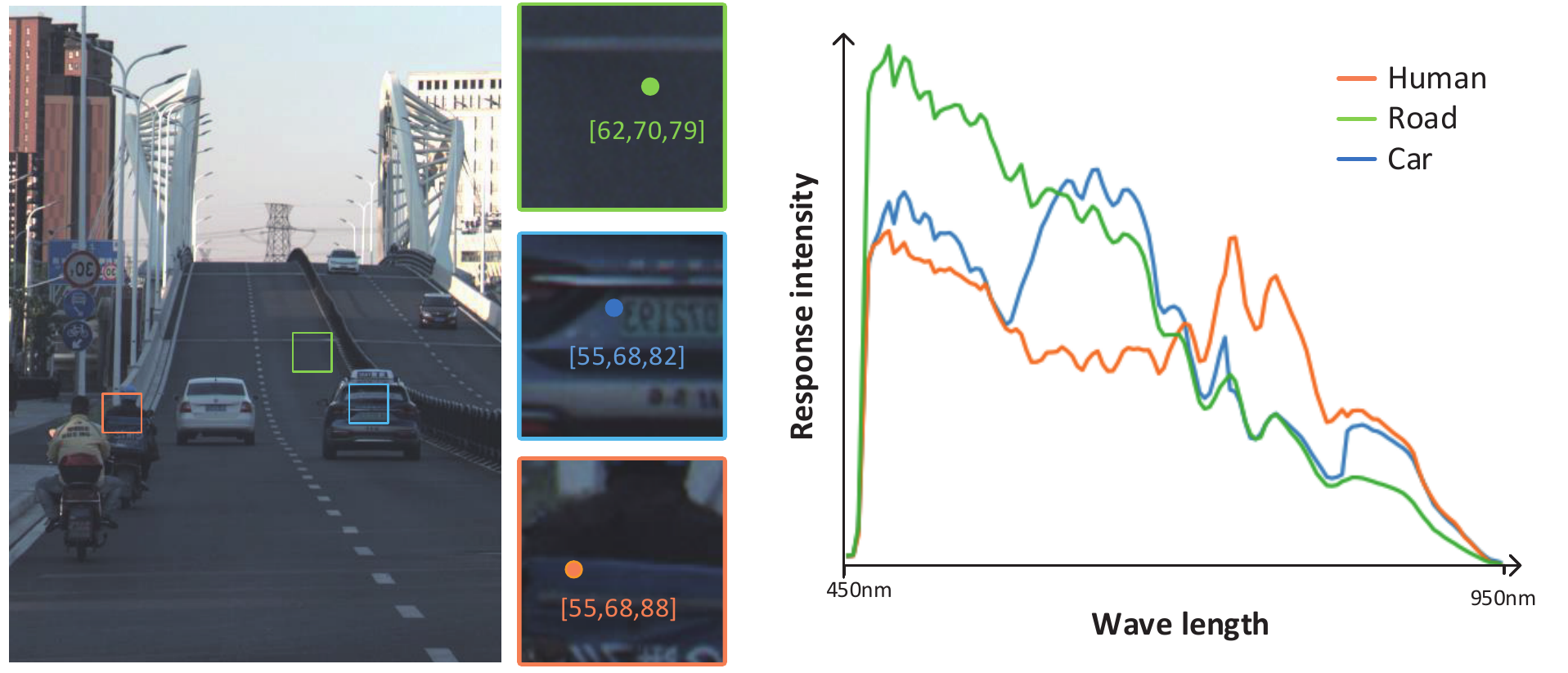}
				\end{minipage}
			}
			\hfill
			\subfloat[Coarse label]{
				\begin{minipage}[b]{0.3\linewidth} 
					\centering
					\includegraphics[width=\linewidth]{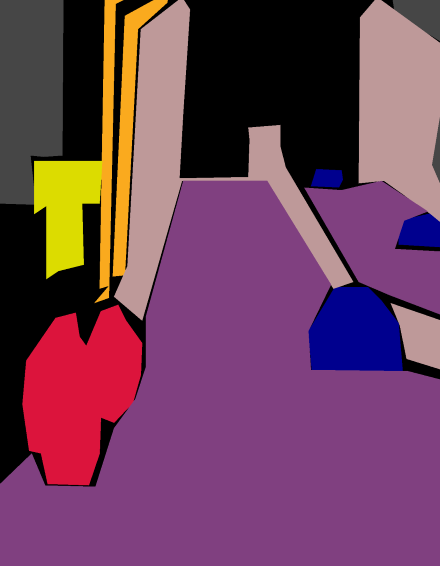}
				\end{minipage}
			}
			\hfill
			\subfloat[HSI result]{
				\begin{minipage}[b]{0.3\linewidth} 
					\centering
					\includegraphics[width=\linewidth]{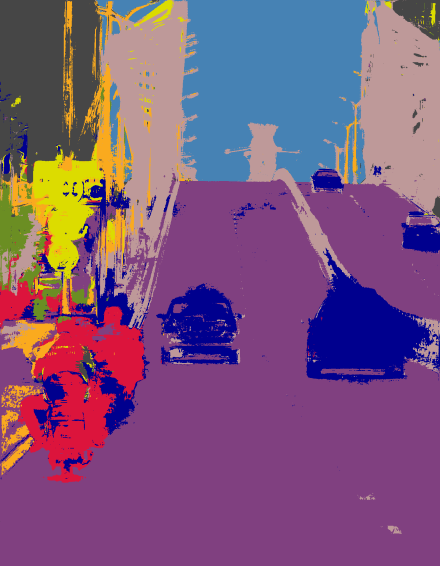}
				\end{minipage}
			}
			\hfill
			\subfloat[RGB result]{
				\begin{minipage}[b]{0.3\linewidth} 
					\centering
					\includegraphics[width=\linewidth]{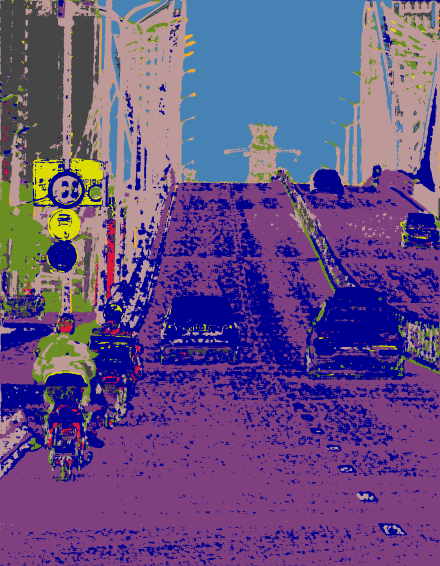}
				\end{minipage}
			}
		\end{minipage}
		\vspace{-12pt}
		\caption{(a) Metamerism: pixels of similar [R,G,B] values may actually have significantly different spectrum. (b) Coarse label. (c, d) An example of spectral classification result based on RGB image and HSI respectively. Based on RGB information, pixels in (a) are all classified as car, but spectrum can distinguish them well.}
		\label{fig:dataset}
		\vspace{-18pt}
	\end{figure}

	We notice, however, most of the RGB based methods relying on large scale and high quality dataset, and/or large and complex networks and fragile training strategies. This is because, RGB images have inherent limitation on metamerism \cite{cao2011high, cao2011prism}. As illustrated in Figure \ref{fig:dataset}(a), different objects may have same RGB value. Metamerism is particularly challenging in cityscape scenes because they contain too many classes, complex lighting and spacial structures.
	%	in cityscape scenes one image contains almost all classes at different scales. 
	%Due to the complexity of the scenes, there are two problems have never been well resolved: (1) A lot of fine manual annotations are needed to provide a prior information, which is expensive, laborious and time-consuming. (2) Feature extraction depends on complex network, which requires a great deal of computing resources. 
	%, which are not discriminative enough in spectral domain especially for the phenomenon of metamerism. New kinds of data with greater capacity are required to break up the status quo.\par
	%The lack of spectral information in RGB images is one of the reasons for these problems. Hyperspectral Image (HSI) captures high resolution spectral and spatial information, which is naturally more informative than RGB images. Figure \ref{fig:dataset}(c,d) gives a comparable results of HSI and RGB image, when classify the pixels mainly based on the values in spectral supervised with coarse label. It is obvious that, in RGB images, substances of similar colors are seriously confused, while the spectrum can distinguish between different semantics well. As shown in Figure \ref{fig:dataset}(a), the spectral response of objects with similar color is very different, a phenomenon known as metamerism. Comparing (c) and (d), the metamerism leads to obvious differences in the classification results of these three points. Because of the insufficient spectral information, vision tasks all heavily depend on the prior between semantics and the spatial structure information. 
	
	In this paper, we propose to introduce HSI into the pipeline of semantic segmentation of cityscape scenes, to break the inherent limitation of RGB images. 
	%Hyperspectral Image (HSI) captures high resolution spectral and spatial information, which is naturally more informative than RGB images. 
	As shown in Figure~\ref{fig:dataset}, while RGB based semantic segmentation easily lead to mistakes if not using large networks, HSI based methods does not have such problem over the same setting.
	%	Beyond human trichromatic vision, HSIs may fundamentally solve the problems faced by visual tasks based on RGB images, which can release the requirements on fine labels and complex networks,\par
	
	Firstly, we theoretically analyze the advantages of HSI in cityscape scenes.
	%we execute comparable and visualized study on HSIs and RGB images used in cityscape scenes, which demonstrates the advantage of HSI theoretically. 
	Specifically, we use t-SNE to show that HSIs are inherently more distinctive than RGB images in cityscape scenes.
	
	Secondly, based on such advantages, we found it is unnecessary to use a fully-supervised network for HSIs, but can still have reliable semantic segmentation. Therefore, we designed a weakly-supervised semantic segmentation framework to further proof the superiority of HSI. 
	Although HSI has inherent advantages, applying it in our task is not trivial. Specifically, a prior relationship between hyperspectral information and semantic categories is established based on coarse label, using our hyperspectral semantic prior module. Additionally, we propose semantic fusion module which fuses the hyperspectral prior with coarse spatial prior. Finally, the finetuning module optimizes the final semantic segmentation output using both spectral prior and spatial prior. For finetuning module, we use the HRNet network \cite{wang2019deep} and DeeplabV3+ \cite{chen2018encoder} in our implementation. However, notice that our proposed framework can generally adopt any RGB segmentation network.

	Finally, we conduct rigorous experiments to demonstrate our proposed idea.
	%Experimental results demonstrate that adopting HSI into cityscape scenes can effectively improve the annotation accuracy 
	Specifically, we conduct experiments on Hyperspectral City V1.0 dataset and found that using coarse labeling and HSI can out perform state-of-the-art RGB methods by $5.0\%$ on mIoU over the same setting.
	%(mIoU) by $5.0\%$ over the state-of-the-art coarse label refinement methods based on RGB images. 
	Furthermore, by using our proposed fine-tuning. We find HRNet \cite{wang2019deep} with the refined labels can improve $10.10\%$ mIoU over that pre-trained on Cityscapes, and $2.16\%$ mIoU over that fine-tuning with coarse label.
	%	Experimental results demonstrate that adopting HSI into cityscape scenes can effectively improve the annotation performance. Our refined label can make $5.0\%$ mIoU improvement over the state-of-the-art coarse label refinement method. For fine-tuning RGB segmentation network, our framework can improve $10.10\%$ mIoU for the HRNet pre-trained on Cityscapes \cite{cordts2016the}, and make $1.91\%$ Acc. and $2.16\%$ mIoU improvement over fine-tuning with coarse label. \par
	
	Our main contributions are summarized as follows:
	\begin{itemize}
		\vspace{-6pt}
		\item To the best of our knowledge, our proposed framework is the first paper to apply HSIs into semantic segmentation in cityscape scenes.
		%To the best of our knowledge, our proposed framework is the first HSIs semantic segmentation framework working with coarse labeling.
		%Aiming at some problems existing in RGB images, we introduce HSIs into semantic segmentation in urban automatic driving scenes and effectively apply the feature of HSIs to semantic segmentation.
		\vspace{-6pt}
		\item We theoretically analyze the necessity of HSI in cityscape scenes. 
		\vspace{-6pt}
		\item We propose a novel weakly-supervised semantic segmentation framework via HSI only works on coarse labels, which is applicable to any RGB semantic segmentation network.
		%We propose a novel weakly-supervised HSI semantic segmentation framework with associate coarse labeled semantics to hyperspectral information and achieve better performance compared with existing method.
		%By extracting the prior relationship between hyperspectral information and coarse labels, we avoid the interference of coarse spatial information and generate high precision refined labels.
		%\vspace{-6pt}
		%\item We cite a novel HSI cityscape semantic segmentation dataset which includes complex cityscapes and lighting conditions.
		%Based on the characteristics of high internal confidence of manual coarse label and high edge fineness of refined label, we propose a label fusion method to combine these two labels. The resulting fusion labels improves the error and precision of coarse labels.
		\vspace{-6pt}
		\item We demonstrate the significant performance improved by adopting HSI to label refinement and semantic segmentation in cityscape scenes, which will likely enable a new research area.
		%We use refined labels to improve the migration performance of the state-of-the-art semantic segmentation model. Our method can significantly improve the segmentation results and retain the segmentation accuracy of the pre-trained model compared with the coarse label.
	\end{itemize}
	
	%-------------------------------------------------------------------------
	\section{Related Work}
	\noindent\textbf{Hyperspectral Image.}
	%Although semantic segmentation using RGB images are well exploited, we find there are few research based on hyperspectral images,
	Hyperspectral images (HSIs) capture the spectral behavior of every pixel within observed scenes at hundreds of continuous and narrow bands, which provides greater information about the captured scenes and objects. HSIs can overcome adverse environmental conditions (\emph{e.g}\onedot, nighttime, foggy, snowy) and reduce the interference of metamerism phenomenon. Several studies \cite{zhou2020improving, lu2020hsi, zhang2021central, xiong2021mcnet} have shown the great potential of spectra in cityscape scenes.\par
	%despite the development of hyperspectral cameras \cite{goetz1985imaging,landgrebe2002hyperspectral,shaw2002signal} and object classification.
	In the past, spectra images are usually acquired by scanning or interferometry in remote sensing (RS)(\emph{e.g}\onedot, Indian Pines~\cite{aviris2012indiana}, Pavia University~\cite{Pavia} and Houston~\cite{debes2014hyperspectral}). But these approaches can only be applied in practice on static or slow-moving scenes. With the advances in compressive sensing theory, snapshot multispectral cameras (\emph{e.g}\onedot, CTIS \cite{descour1995computed}, PMVIS \cite{cao2011high} and SPCS \cite{august2013compressive}) can measure data in a single exposure on sensor. At present, the acquisition technology has been able to capture high-resolution spectral video \cite{chen2017high-resolution}, which greatly expands the application field of spectral imaging. \par

	\noindent\textbf{Semantic Segmantation in Cityscape Scenes.}
	Semantic segmentation is a task of predicting unique semantic label for each pixel of the input image. It has achieved great progress with the works such as FCN~\cite{Long_2015_CVPR}, UNet~\cite{ronneberger2015u-net:}, SegNet~\cite{badrinarayanan2017segnet:}, PSPNet~\cite{zhao2017pyramid}, DeepLabv3~\cite{chen2017rethinking} and HRNet~\cite{wang2019deep}. 
	%And the extended receptive field is used to extract larger neighborhood and multi-scale spatial information \emph{e.g}\onedot, ASPP~\cite{chen2017rethinking}, PPM~\cite{zhao2017pyramid}. Some methods establish long distance context relationships between pixels \emph{e.g}\onedot, CRF~\cite{chandra2016fast,krahenbuhl2011efficient} and attention~\cite{xue2019danet,yuan2018ocnet,yuan2019object}. \par
	Fully-supervised Semantic Segmentation depending on huge datasets with pixel-wise annotation (\emph{e.g}\onedot, Cityscapes~\cite{cordts2016the}, KITTI \cite{geiger2012we} and CamVid~\cite{brostow2009semantic}) is expensive and labor-consuming. 
	%The core is to extract spatial information and generate high-level features through deep residual network. This leads to two problems:(1) Spatial information relies on careful manual annotation, which requires a large amount of labor and time cost.(2) Large segmentation networks require large computational cost trained from scratch. So we hope to solve these problems by introducing new spectral information from HSIs.
	
	To solve this problem, numerous papers focus on semi- and weakly-supervised semantic segmentation. The semi-supervised methods, such as video label propagation \cite{chennaive, mustikovela2016can,budvytis2017large}, consistency regularization \cite{french2019semi-supervised, mittal2019semi-supervised}, self-training \cite{luc2017predicting, zou2018unsupervised, lian2019constructing, li2019bidirectional, zou2019confidence} have made great effect, but still rely on the fine annotations. Weakly-supervised methods usually employ bounding boxes \cite{dai2015boxsup}, scribbles \cite{lin2016scribblesup}, points \cite{bearman2016s} and image-level labels \cite{papandreou2015weakly}. For image-level labels, most of methods \cite{ahn2018learning, hou2018self, wei2017object} refine the class activation map (CAM) \cite{zhou2016learning} generated by the classification network to approximate the segmentation mask. Besides, the network also be trained with bounding boxes \cite{khoreva2017simple, papandreou2015weakly}, scribbles \cite{lin2016scribblesup}, or videos \cite{saleh2017bringing}. But comes to complex cityscape scenes, it is difficult to utilize these labels for one image contains almost all classes. For cityscape scenes, annotating coarsely only requires each polygon must only include pixels belonging to a single class, which is a low cost weakly-supervised labels \cite{cordts2016the}. But due to the sparse supervision, using coarse label directly can not achieve competitive results. 
	
	%\noindent\textbf{Weakly-supervised Semantic Segmentation.}
	%The acquisition of pixel-level labels in fully-supervised learning is time consuming and laborious, weakly supervised learning is gradually achieving good results at a lower cost. \par 
	%There are numerous papers focus on semi-supervised semantic segmentation, such as video label propagation~\cite{chennaive, mustikovela2016can,budvytis2017large}, consistency regularization~\cite{french2019semi-supervised,mittal2019semi-supervised}, self-training~\cite{luc2017predicting, zou2018unsupervised,lian2019constructing,li2019bidirectional,zou2019confidence}. However, all of these methods are semi-supervised with the help of some fine annotations.
	
	\noindent\textbf{Annotation Refinement.}
	There are rare methods focusing on the refinement of coarse labels. In \cite{yang2016object}, a fully convolution encoder-decoder network with the dense conditional random field (CRF) is proposed for contour detection in order to refine imperfect annotations. However, the large computational cost and sensitivity to parameter selection restrict its practicability. \cite{luo2018coarse-to-fine} proposes a coarse-to-fine annotation enrichment strategy which expends coarse annotations to a finer scale. But coding and iterating also make it too complex. Fundamentally, in the absence of the prior between spatial information and semantics in coarse labels, it is difficult to refine coarse labels directly.
	
	%-------------------------------------------------------------------------
	\section{Theoretical Analysis}
	%Traditional data sets based on RGB images mainly contain spatial information and spectral information only has RGB three channels. The disadvantages of data sets also limit the development of technology. In recent years, with the rapid development in computational photography theory and semiconductor techniques, spectral video acquisition has become feasible~\cite{chen2017high-resolution}. Since each material has unique signatures in a spectral domain, spectral imaging can provide much more information about the captured scenes. New datasets based on space-spectrum information are necessary. Hyperspectral city dataset is an urban autonomous driving scene spectral dataset which utilizes newly developed hyperspectral camera. The inclusion of both spatial and spectral information is of great significance for developing a new direction of semantic segmentation.
	
	To enable our research in HSIs for low cost and reliable semantic segmentation in cityscape, we adopt a new dataset called the Hyperspectral City Dataset. Based on this dataset, we will introduce the great advantages HSIs present over RGB images in the cityscape scenes.
	
	%Spectral information is influenced by spectral resolution. Spatial information is influenced by spatial resolution and shooting scenes (cityscape scenes contain more complex object information than aerial scenes).
	
\begin{figure}[t]
		\centering
		\subfloat[Hyperspectral City V1.0]{
			\begin{minipage}[b]{\linewidth} 
				\centering
				\includegraphics[width=\linewidth]{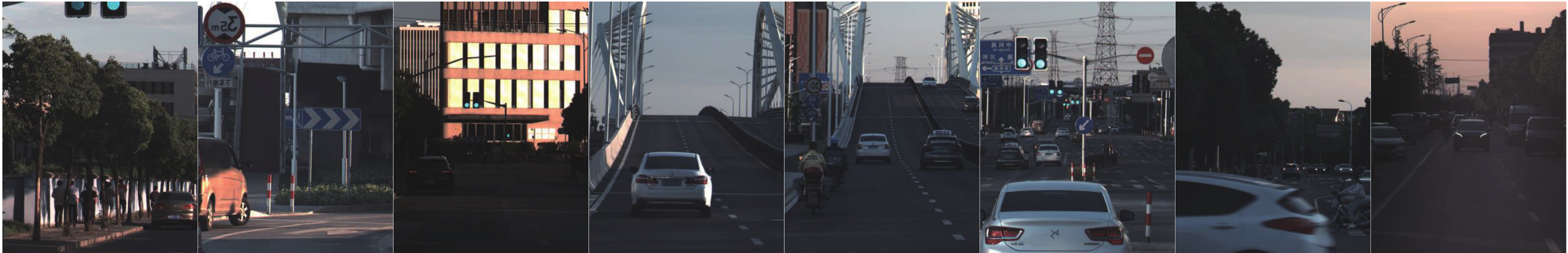}
			\end{minipage}
		}
		
		\subfloat[Cityscapes]{
			\begin{minipage}[b]{\linewidth} 
				\centering
				\includegraphics[width=\linewidth]{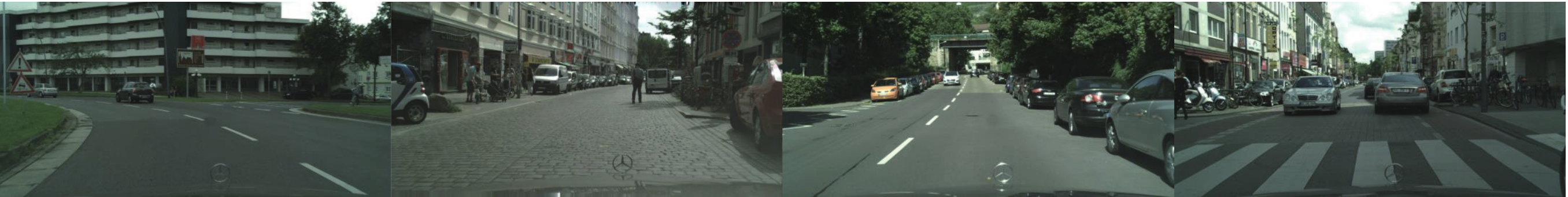}
			\end{minipage}
		}
		
		\subfloat[KITTI]{
			\begin{minipage}[b]{\linewidth} 
				\centering
				\includegraphics[width=\linewidth]{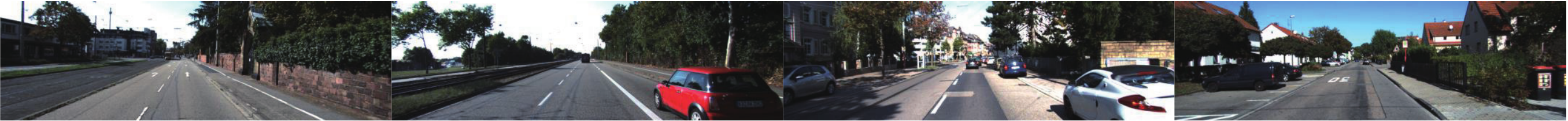}
			\end{minipage}
		}
		
		\subfloat[Comparison of RGB and hyperspectral datasets]{
			\begin{minipage}[b]{\linewidth} 
				\centering
				\includegraphics[width=\linewidth]{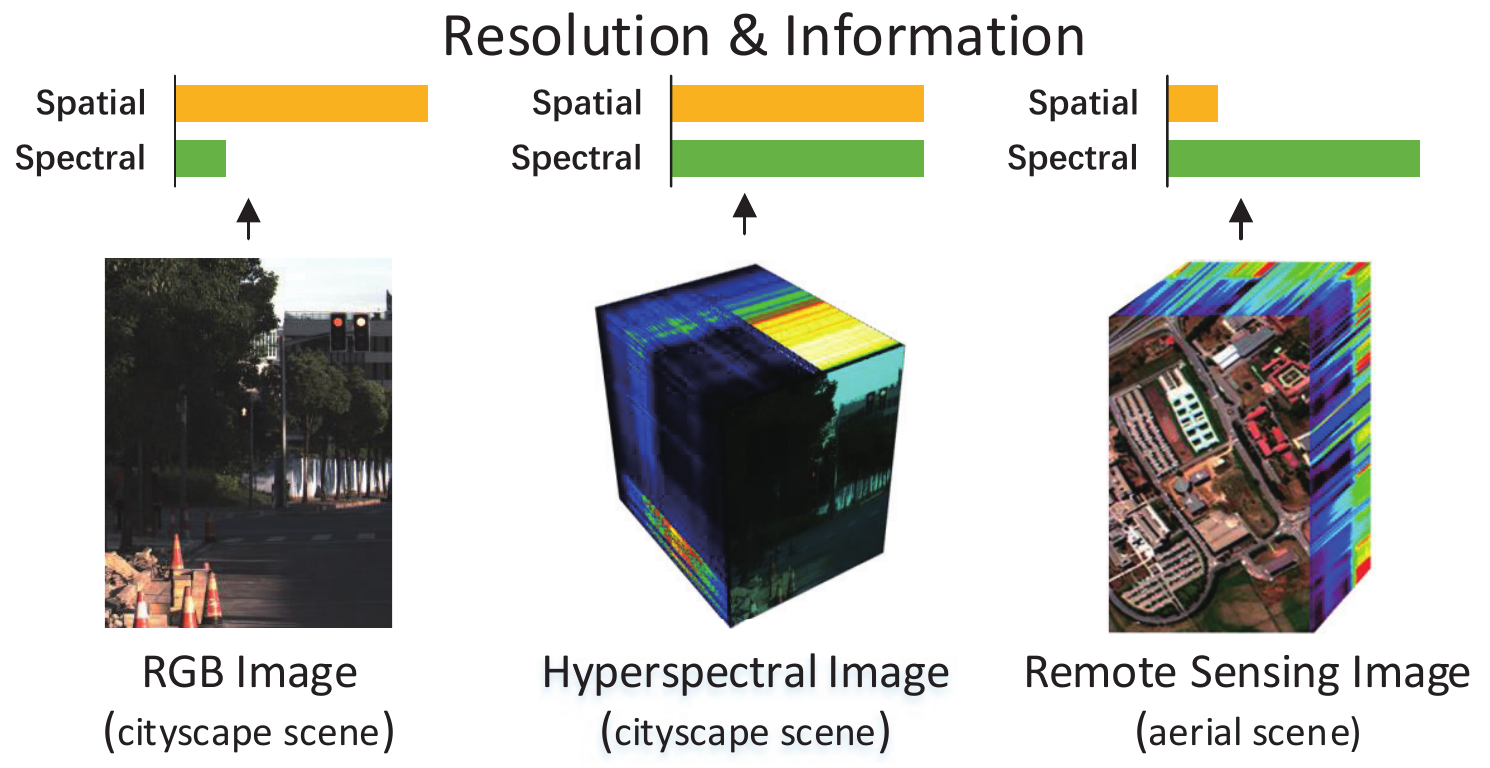}
			\end{minipage}
		}
		\caption{Comparisons of Hyperspectral City dataset with other datasets: Examples from (a)  Hyperspectral City V1.0, (b) Cityscapes~\cite{cordts2016the}, (c) KITTI~\cite{geiger2012we}. (d) Comparison of RGB images, HSIs and remote sensing images. The orange and green bars represent roughly the spectral and spatial resolution of the image, as well as the corresponding amount of information. }
		\vspace{-14pt}
		\label{fig:dataset image}
	\end{figure}

	To best exploit the feasibility of modern semantic segmentation, the dataset focuses particularly on complex cityscape scenes as well as complex lighting conditions. As Figure~\ref{fig:dataset image} shows, compared with other cityscape scenes datasets (\emph{e.g}\onedot, Cityscapes~\cite{cordts2016the} and KITTI~\cite{geiger2012we}), the scenes and weather conditions in this dataset are more complicated.
	
	\noindent\textbf{HSI Acquisition.}
	While HSI have been exploited in remote sensing, it cannot be directly used in cityscapes. 
	Remote sensing images are acquired based on scanning or interferometry methods \cite{gat2000imaging, morris1994imaging}, which limit the use of spectra. The Hyperspectral City Dataset was captured by PMVIS \cite{cao2011high}, which can capture both high-resolution spectral and RGB videos in real time. Specifically, as shown in Figure \ref{fig:dataset image}, the PMVIS camera works well in highly dynamic and complex scenes such as cityscape.
	%This greatly expands the spectrum application scenes. Therefore, HSIs in cityscape scenes have essential differences between remote sensing images in scenes and imaging methods.  

	\begin{figure}[t]
		\centering
		\begin{minipage}[b]{\linewidth}
			\subfloat[Hyperspectral image]{
				\begin{minipage}[b]{0.5\linewidth} 
					\centering
					\includegraphics[width=\linewidth]{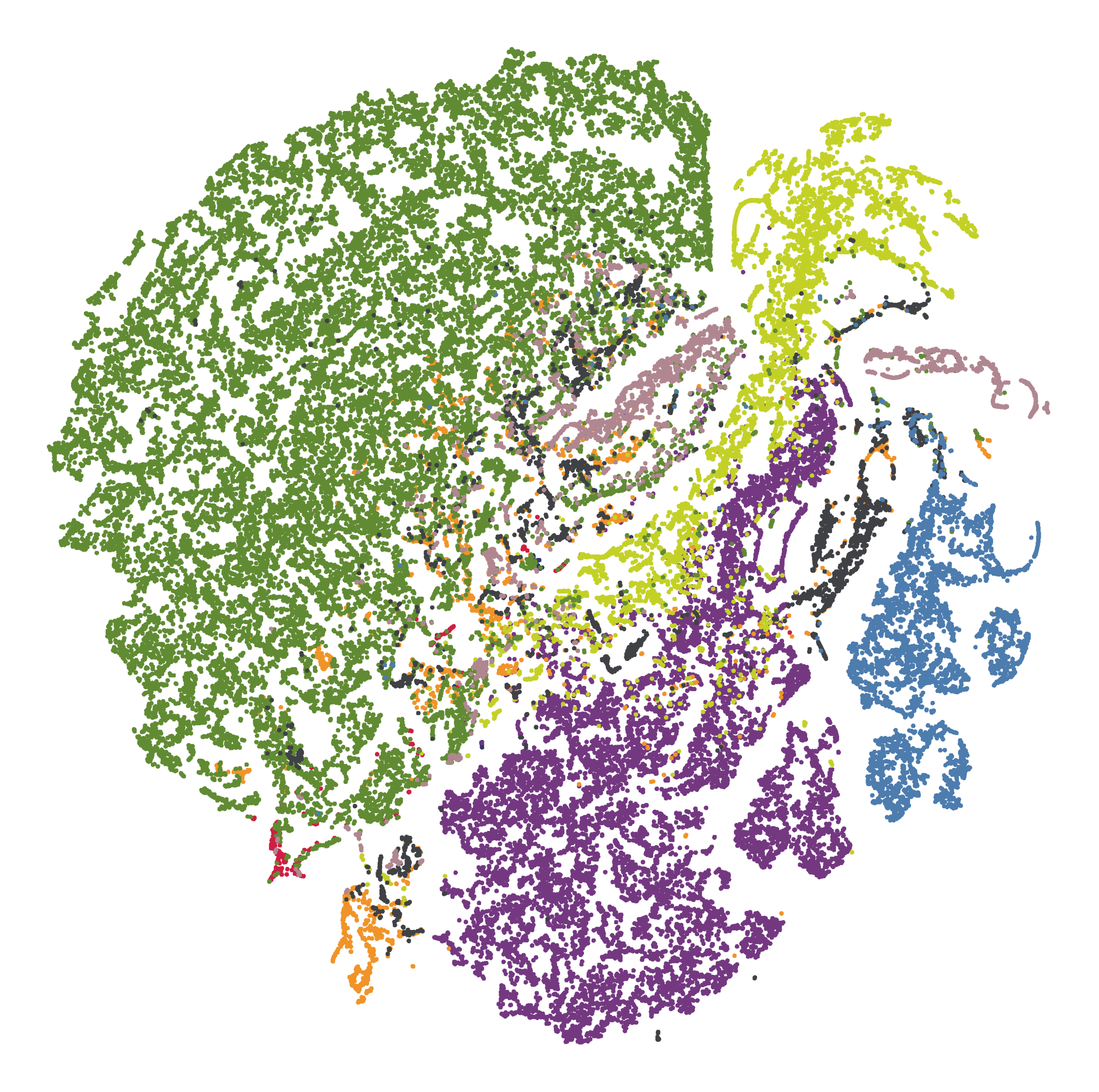}
				\end{minipage}
			}
			\subfloat[RGB image]{
				\begin{minipage}[b]{0.5\linewidth} 
					\centering
					\includegraphics[width=\linewidth]{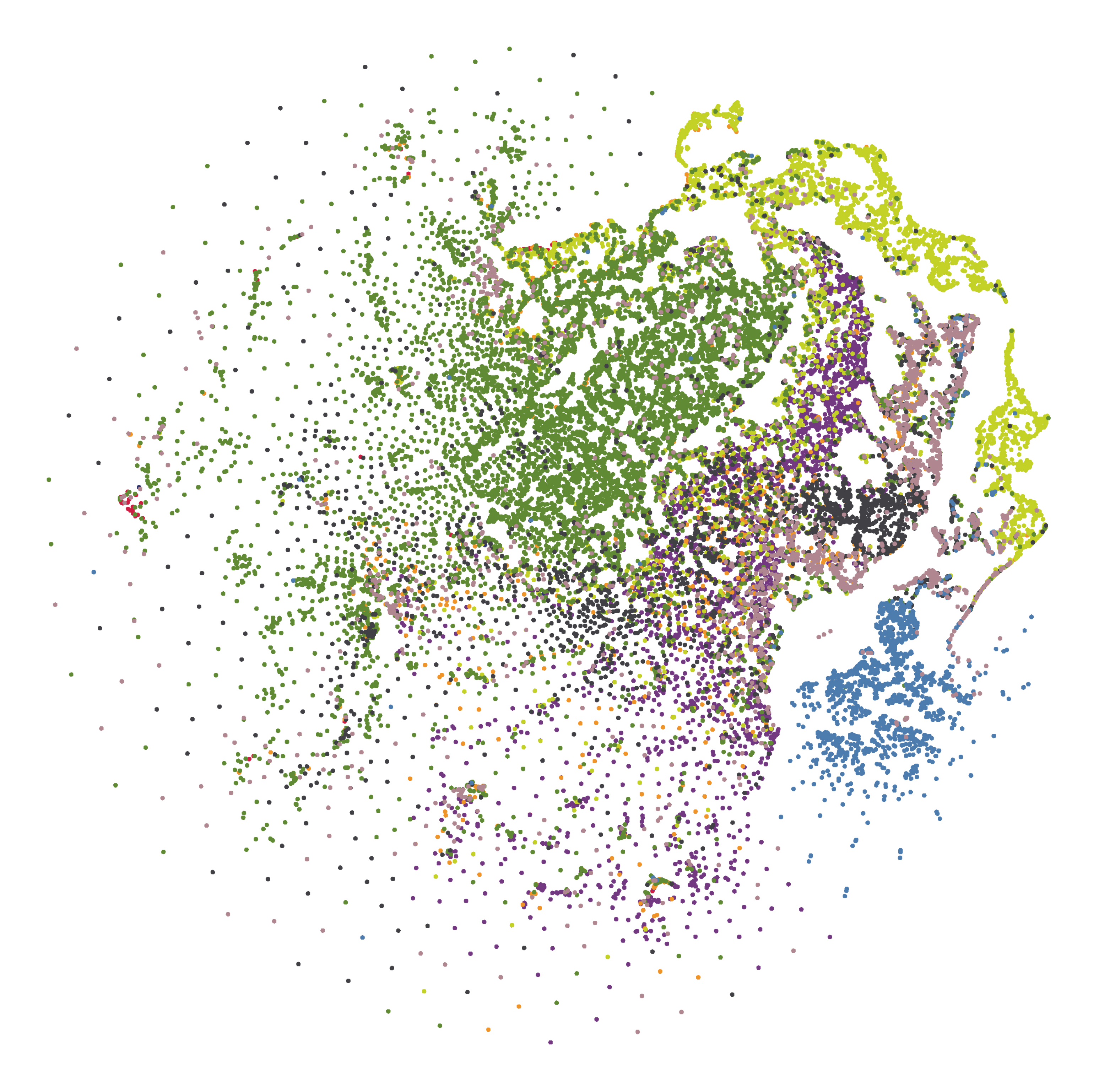}
				\end{minipage}
			}
			\vfill
			\includegraphics[width=\linewidth]{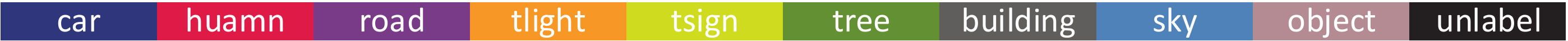}
		\end{minipage}
		\caption{Visualizing data using t-SNE. We performed t-SNE visualization of RGB images and HSIs respectively under the same experimental conditions. Compared with RGB images, HSIs have stronger category prior.}
		\label{fig:TSNE}
		\vspace{-18pt}
	\end{figure}
	
	\noindent\textbf{HSI in natural scenes and difference from remote sensing.}
	As shown in Figure \ref{fig:dataset image} (d), We compare three kinds of data. The middle cube represents the HSI in Hyperspectral City, which has high spatial and spectral resolution. The left cube represents the RGB image, which is the result of the integration of the spectral image over the spectral channel. A lot of spectral information is lost in the process of integration, therefore RGB image mainly contains spatial information. The right cube represents the aerial remote sensing image, which has high spectral resolution and low spatial resolution. The different imaging scenes and acquisition methods make a great difference between cityscape HSI and remote sensing image.\par% The imaging scenes and methods are quite different from urban scene datasets. %Figure \ref{fig:dataset image} illustrates that the HSIs of urban scene have abundant both spatial and spectral information.
	
	\noindent\textbf{Hyperspectral information as semantic feature.}
	Hyperspectral images have higher spectral resolution, so they have better discrimination of semantic features. To prove that each pixel of HSIs has stronger semantic properties than RGB images, we use t-SNE ~\cite{maaten2008visualizing} to visualize HSI and RGB image. We select the HSI and RGB image in the same frame, which have the same fine label. Due to computational limitations, we use the nearest interpolation to sample the HSI, the RGB image and the fine label to a same low resolution, which keeps each pixel corresponding. We use HSI and RGB image respectively to create a t-SNE visualization. As shown in Figure \ref{fig:TSNE}, the result of HSI has a continuous distribution and a clear boundary between each category. On the contrary, for RGB image, the confusion of spectra is more serious.
	%The same semantic categories contain similar categories of objects, and the spectrum can accurately reflect the characteristics of these categories. With the exclusion of spatial structure information, the t-SNE visualization result illustrates that HSIs have richer semantic features than RGB images. Compared with general visual scene, the urban automatic driving scene contains multi-scale and more complex ground objects, the semantic information of hyperspectral pixels can well deal with the challenges of multi-scale and complex scenes. Therefore, hyperspectral images have a good application value in semantic segmentation. 
	Figure \ref{fig:dataset}(a) plots the spectral curves of different substances with the similar color. HSI can reduce the interference of metamerism phenomenon and provide more powerful information support for cityscape scene analysis.
	
	%-------------------------------------------------------------------------
	\section{Weakly-supervised HSI Semantic Segmentation Framework}
	
	\begin{figure*}
		\centering
		\includegraphics[width=1\linewidth]{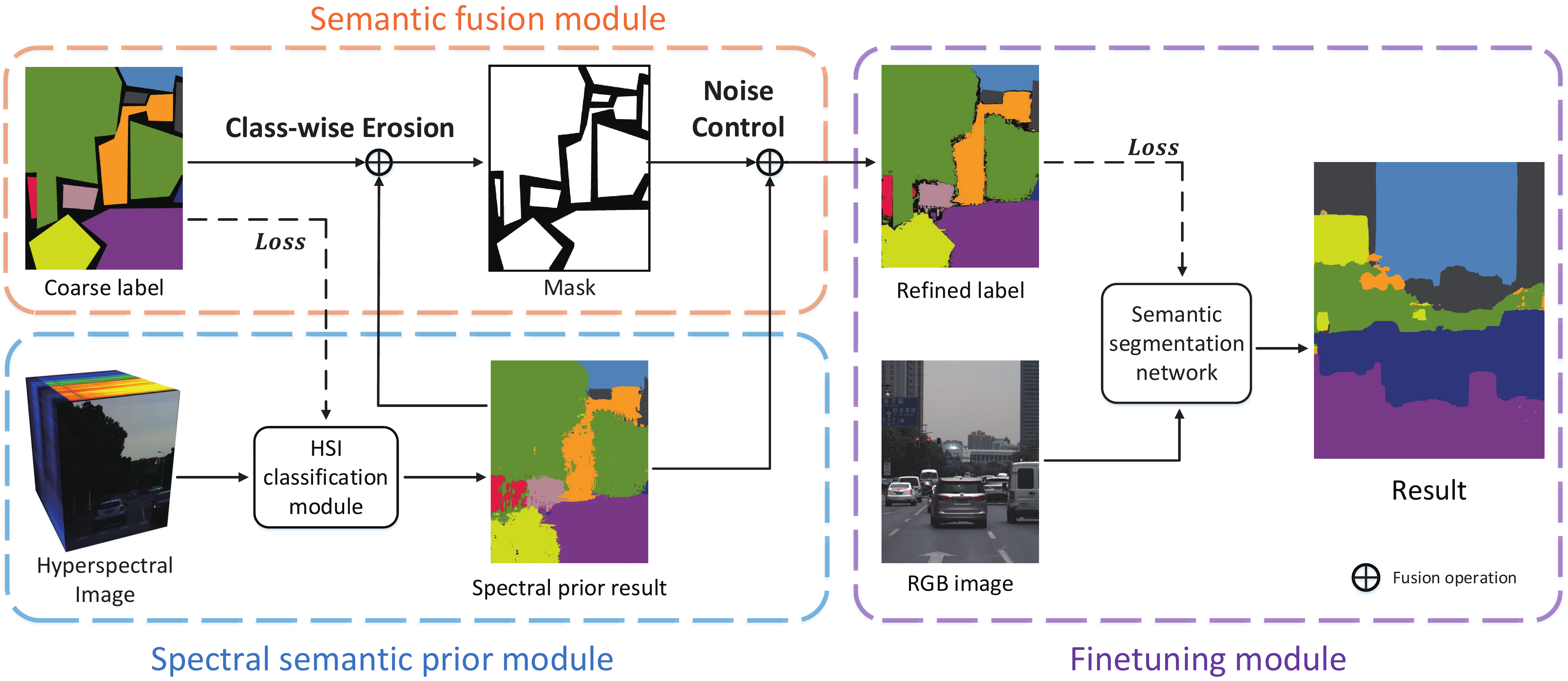}
		\vspace{-12pt}
		\caption{The proposed framework. In the \textcolor[RGB]{0,0,255}{blue} dashed box, we train HSI classification module with coarse label as supervision, and then input HSI to generate spectral prior result. In the \textcolor[RGB]{237,125,49}{orange} dashed box, we combine the coarse label and the spectral prior through the mask to generate the refined label. In the \textcolor[RGB]{112,48,160}{purple} dashed box, we fine-tune the HRNet pre-trained model with the refined label as the supervision.}
		\vspace{-12pt}
		\label{fig:framework}
	\end{figure*}
	
	\begin{figure}
		\centering
		\includegraphics[width=1\linewidth]{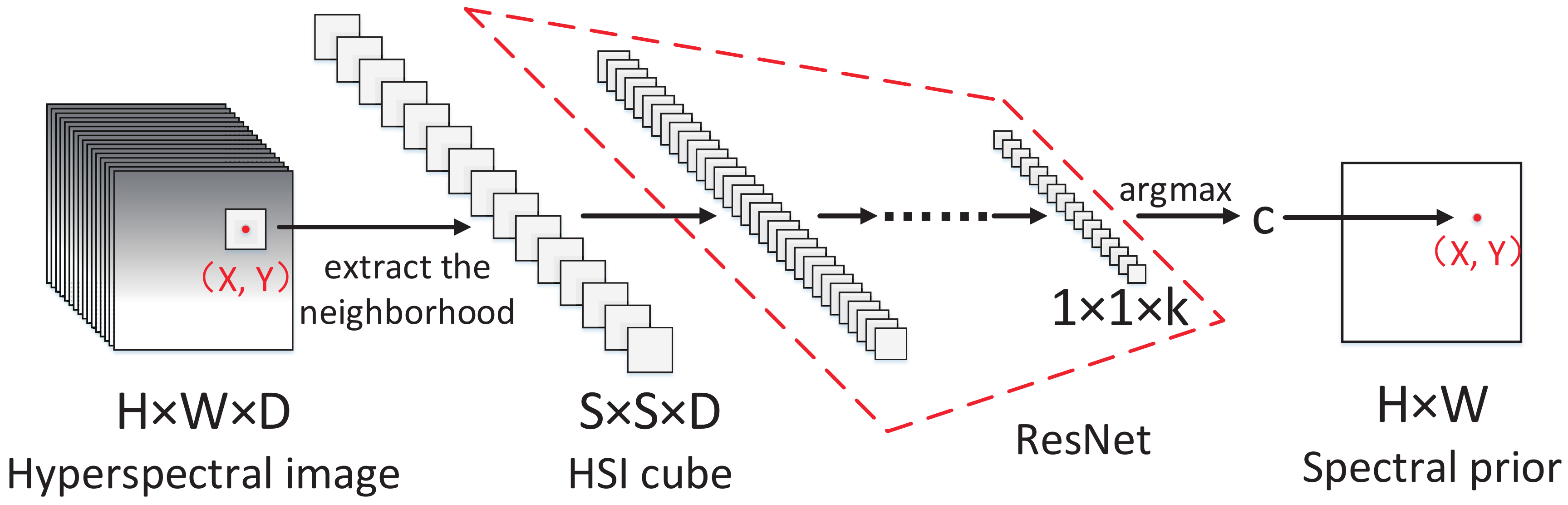}
		\vspace{-18pt}
		\caption{The illustration of HSI classification module. ResNet50 is used to classify the HSI cubes and obtain the label of the pixel. For each pixel of the HSI is classified and finally generate a spectral prior result.}
		\label{fig:hsiclassification}
		\vspace{-18pt}
	\end{figure}
	%\shaodi{Name your method}
	
	%In this section, we introduce our HSI based semi-supervised semantic segmentation network.
	%based on HSIs rich in spectral information, a hyperspectral refinement framework on semantic segmentation is described in detail, including refined label generation, label fusion and training semantic segmentation network. 
	
	\subsection{Overview}
	%HSIs have a higher number of channels than RGB images, and each pixel contains higher resolution spectral information (visible light and near infrared). Spectral information reflects the response of the object at different wavelengths. In this sense, HSIs of cityscapes scene contain rich spectral information and spatial information, and these information complement each other. For example, on the one hand, in the case of occlusion, spectral information can make up for the lack of spatial information. On the other hand, in the complex category of internal matter, spatial information can be a good complement to spectral information.\par
	
	%Although HSIs are rich in information, they also have some defects, such as memory occupation and large computation. And due to the limitation of image size, the number of images in the spectral dataset is much smaller than that in the RGB dataset. 
	Although HSI has inherit advantages, applying it in our task in not trivial.
	A hyperspectral image requires more than 1G memory. It is problematic that the naive extension of existing network structure by just increasing the channels with research in memory overflow. From the perspective of label, our method avoids the limitation of memory, effectively reduces the cost of labeling and improves the performance of segmentation.
	
	%Therefore, it is difficult to use HSIs directly in semantic segmentation. 
	%We use spectral information to optimize labels by using the complementary properties of spatial information and spectral information. While making full use of the existing mature semantic segmentation of RGB images, the advantages of HSIs are also utilized.\par
	
	As illustrated in Figure \ref{fig:framework}, there are 3 modules of our method. 
	%\shaodi{Update this paragraph based on my modification of the introduction and make it consistent with my modification.}
	First, coarse label is used to supervise HSI classification to generate a spectral prior result. This step gets a prior relationship between HSI and coarse label to generate high edge precision result. Second, coarse label and spectral prior are fused to generate more detailed and accurate label, which is called refined label. Third, the refined label is utilized as a supervision to improve the migration effect of the mature semantic segmentation pre-trained model. \par
	Formally, given a set of M training data $X_h, X_r, Y = \{I_i\}^M_{i=1}$, let $\bold X_h \in R^{H\times W \times D_h}$ and $\bold X_r \in R^{H\times W \times D_r}$ denote a pair of hyperspectral image data and RGB image data, where $H$ and $W$ are the spatial dimensions of the input tensor, height and width, and $D_h$, $D_r$ is the number of spectral channels. Every $X_h, X_r$ has a pixel at location $x,y$ contains a same one-hot label $\bold Y_{x,y} = (y_1, y_2,...,y_k) \in R^{1\times 1\times k}$ where $k$ represents the number of classes.\par

	%-------------------------------------------------------------------------
	\subsection{Hyperspectral Semantic Prior Module}
	In this section, we use HSI classification module to generate semantic prior. 
	We hope to use the prior relationship between coarse label and hyperspectral information to obtain label with high fineness, and at the same time, to prevent the influence of coarse spatial information of coarse label.
	%Based on hyperspectral classification, we propose a semantic segmentation label generation method with high precision.  \par
	
	As shown in Figure \ref{fig:hsiclassification}, we first generate HSI cube $C \in R^{S\times S\times D_h}$ with the size $S\times S$ from $X_h$, whose center at the space position is $(x,y)$ where $x\in[(S-1)/2+1, H-(S-1)/2]$, $y\in[(S-1)/2+1, W-(S-1)/2]$. Thus, a HSI cube at the position $(x,y)$ is denoted by $C_{x,y}$. The HSI cube covers the height from $x-(S-1)/2$ to $x+(S-1)/2$, the width from $y-(S-1)/2$ to $y+(S-1)/2$ and the whole spectral dim $D_h$. The number of HSI cubes generated from $X_h$ is $(H-S+1)\times (W-S+1)$. The label of the $C_{x,y}$ is the one-hot label $Y_{x,y}$ of the pixel at the position $(x,y)$.\par
	After we generate HSI cubes from the dataset, we use ResNet-50~\cite{he2016deep} as the hyperspectral classification network. We change the input dims of ResNet50 from $D_r$ to $D_h$ to adapt HSI cube. During training, we learn the label under the supervision from the ground-truth $Y_{x,y}(c) = 1 c\in[0,k]$ using the cross-entropy loss and $f(C_{x,y}) = Z_{x,y}$ the output of ResNet50, as the following equation shows:
	\begin{equation}
		\setlength{\abovedisplayskip}{3pt}
		\setlength{\belowdisplayskip}{3pt}
		\begin{aligned}
			L(Z_{x,y}, c)& = -\log\left(\frac{\exp(Z_{x,y}[c])}{\sum_k\exp(Z_{x,y}[k])}\right) \\
			&= -Z_{x,y}[c] + \log\left(\sum_k \exp(Z_{x,y}[k])\right) \ . \label{0}
		\end{aligned}
	\end{equation}
	After training, we use the hyperspectral classification network to compute the result for each pixel of HSI $X_h$ to generate spectral prior $Z$. \par
	\subsection{Semantic Fusion module}
	In this section, we combine coarse labels and spectral prior to generate refined labels. Although spectral prior $Z$ have higher edge fineness, manual coarse labels have higher confidence in the central region. So a label fusion algorithm is proposed to fuse the advantages of two kinds of labels. First, we remove the low confidence pixels in the spectral prior. Then, we use a class-based erosion strategy to combine spectral prior $Z$ and coarse label $Y$ to generate refined label. \par
	\noindent\textbf{Noise control.} For each pixel $Z_{x,y}$ in spectral prior, We use the softmax function to calculate the confidence, where the softmax function is defined as:
	\begin{equation}
		f_{softmax}(Z_{x,y}) = \frac{\exp(Z_{x,y}[k'])}{\sum_k\exp(Z_{x,y}[k])}
	\end{equation}
	where $k' = argmax(Z_{x,y})\in[1,\,k]$. For pixel $Z_{x,y}$, if the confidence is below the threshold $\alpha$, the pixel will be set to the label of 'background'; otherwise it is assigned the label class-$k'$. It plays an important role in controlling spectral prior quality. \par
	\noindent\textbf{Class-wise erosion fusion.} Due to the edge of manual coarse label will have some errors beyond the boundary of the classes. At the same time, in the internal area of some classes (\emph{e.g}\onedot, car, building), spectral prior has misclassification. So we propose a class-wise erosion kernel size selection method to obtain the optimal mask, and then fuse two labels.\par 
	Coarse label $Y\in R^{1\times 1\times k}$ is a one-hot label. For each class $ks$, the coarse label $Y_k$ is eroded by each category. We choose a square $E$ with size $l\times l$ as the kernel of erosion. For class $i \in [1,\,k]$, the erosion operation as the following equation shows:
	\begin{equation}
		\setlength{\abovedisplayskip}{3pt}
		\setlength{\belowdisplayskip}{3pt}
		f_{erode}({Y_i} (x,\,y)) =  \min _{x',\,y'\in (-l, \,l),  \, l \ne0 } {Y_i} (x+x',\,y+y') \ . \label{1}
	\end{equation}
	After eroding each class, the regions eroded by each class are added together to form a mask. The mask $Y_{mask}$ after the erosion operation retains the area near the center of each class. Then we use a mask $Y_{mask}$ to fuse the spectral prior $Z$ with the coarse label $Y$ to generate refined label $Y_{refined}$, as the following equation shows:
	\begin{equation}
		\setlength{\abovedisplayskip}{3pt}
		\setlength{\belowdisplayskip}{3pt}
		Y_{refined} = Y \times \sum_{i=1}^{k} f_{erode}(Y_i) + Z \times (1-\sum_{i=1}^{k} f_{erode}(Y_i)) \ . \label{3}
	\end{equation} \par
	Next we use class-wise intersection over union (IoU) as the evaluation index to calculate the IoU scores of each class under different erosion kernel sizes $l \in (1,\,n)$. We select the erosion kernel size $l_i$ with the highest IoU score for each class. By this method, we obtained the final optimal erosion kernel size $l_i$ for each class $i\in[1,\,k]$. Finally, we generate mask with kernel size $l$. \par
	Obviously, refined label combines the high internal confidence of coarse label and the high edge fineness of spectral prior. Using hyperspectral information, we get high quality label only based on coarse label.
	
	\subsection{Finetuning Module}
	The network structure of semantic segmentation has been relatively mature. To make use of the existing semantic segmentation mature network and prove that our method is useful to semantic segmentation, we fine-tune the HRNet and DeeplabV3+ pre-trained model with our refined labels. More details can be found in supplementary material.\par
	
	%In more detail, we use HRNetV2-W48~\cite{sun2019high} pretrained on Cityscapes as baseline. HRNet first uses two strided $3\times 3$ convolutions to decrease the resolution to $1/4$. Then it contains 4 stages that are formed by repeating modularized multi-resolution blocks. The last layer mix the output of the four resolution channels by a $1\times 1$ convolution and the output are upsampled (4 times) to the input size by bilinear upsampling. HRNet augment the high-resolution representation by aggregating the (upsampled) representations from all the parallel convolutions. This leads to stronger representation and get superior results. And in order to prevent overfitting, we fixed the parameters of the first two strided $3\times 3$ convolutional layers and the 4 stages of HRNet. Only the last two $1\times 1$ convolution layers are trained. Although this may weaken the degree of enhancement, we want to prove that our method has a good meaning for semantic segmentation.\par
	
	%-------------------------------------------------------------------------
	
	\section{Experiments}
	
	%^To evaluate the proposed method, we carry out experiments on Hyperspectral City V1.0 dataset. This section, we first introduces the dataset and implementation details, then we perform a series of ablation experiments of label generation and semantic segmentation on Hyperspectral dataset.
	\begin{figure*}[t] 
		\centering 
		\begin{minipage}[b]{\linewidth} 
			\subfloat[Raw RGB]{
				\begin{minipage}[b]{0.12\linewidth} 
					\centering
					\includegraphics[width=\linewidth]{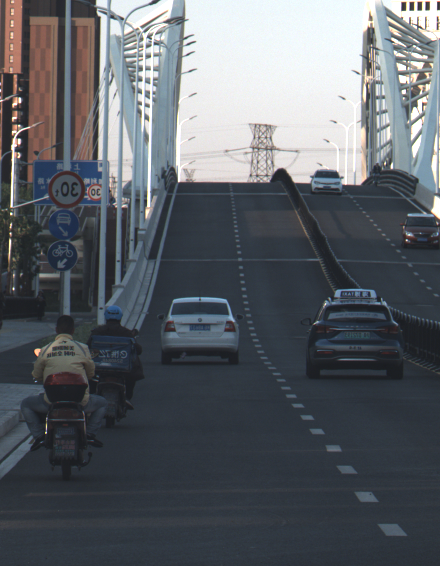}\vspace{4pt}
					\includegraphics[width=\linewidth]{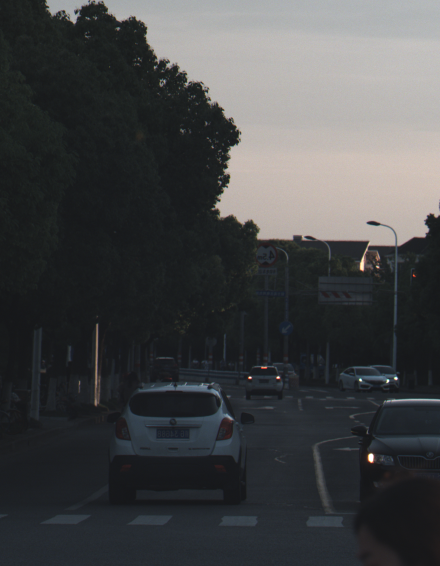}
				\end{minipage}
			}
			\hfill
			\subfloat[Ground-truth]{
				\begin{minipage}[b]{0.12\linewidth}
					\centering
					\includegraphics[width=\linewidth]{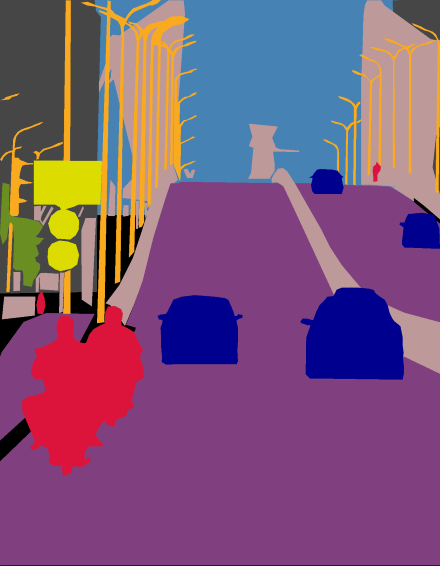}\vspace{4pt}
					\includegraphics[width=\linewidth]{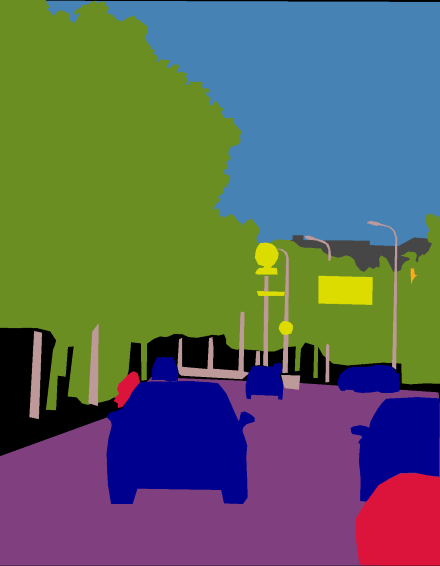}
				\end{minipage}
			}
			\hfill
			\subfloat[Coarse label]{
				\begin{minipage}[b]{0.12\linewidth}
					\centering
					\includegraphics[width=\linewidth]{images/coarse5.png}\vspace{4pt}
					\includegraphics[width=\linewidth]{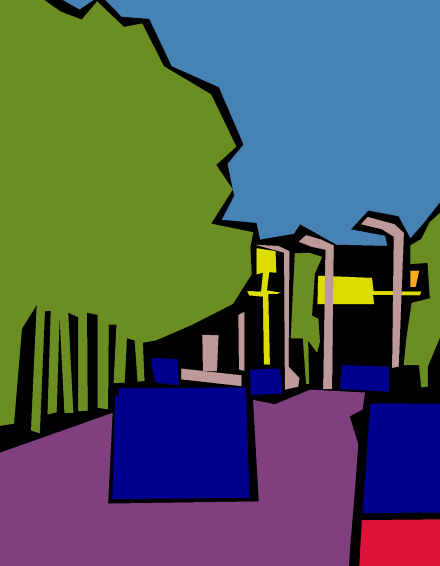}
				\end{minipage}
			}
			\hfill
			\subfloat[\centering Spectral prior (HSI)]{
				\begin{minipage}[b]{0.12\linewidth}
					\centering
					\includegraphics[width=\linewidth]{images/4.1.png}\vspace{4pt}
					\includegraphics[width=\linewidth]{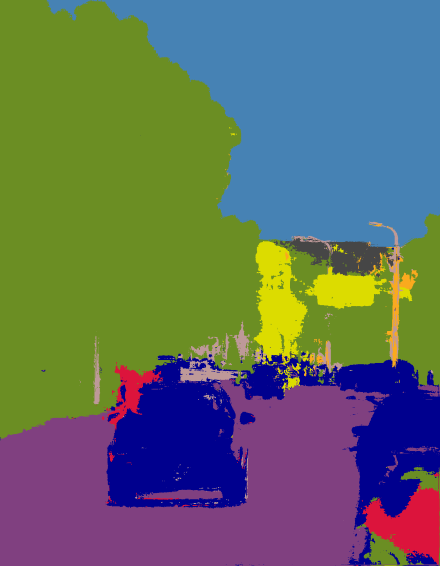}
				\end{minipage}
			}
			\hfill
			\subfloat[\centering Spectral prior (RGB image)]{
				\begin{minipage}[b]{0.12\linewidth}
					\centering
					\includegraphics[width=\linewidth]{images/4.3.png}\vspace{4pt}
					\includegraphics[width=\linewidth]{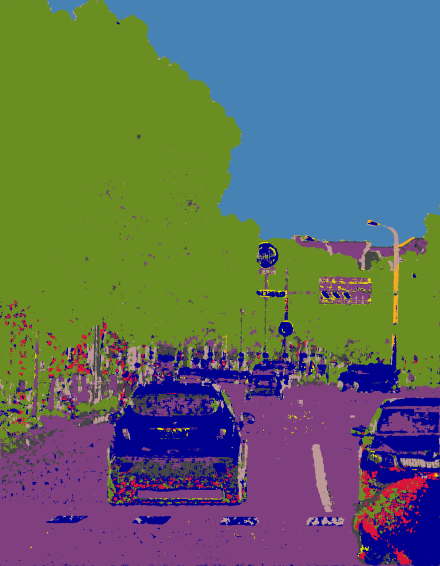}
				\end{minipage}
			}
			\hfill
			\subfloat[\centering Coarse-to-fine(0.7)]{
				\begin{minipage}[b]{0.12\linewidth}
					\centering
					\includegraphics[width=\linewidth]{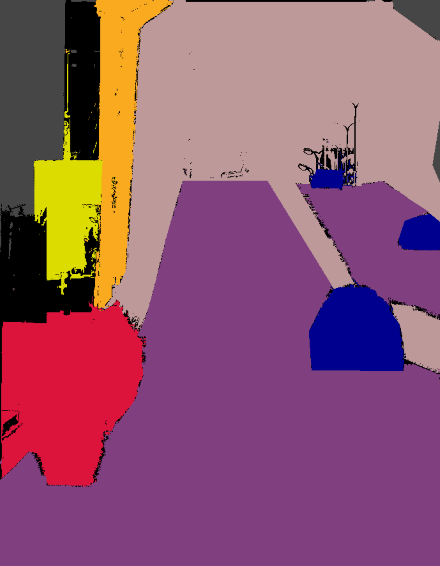}\vspace{4pt}
					\includegraphics[width=\linewidth]{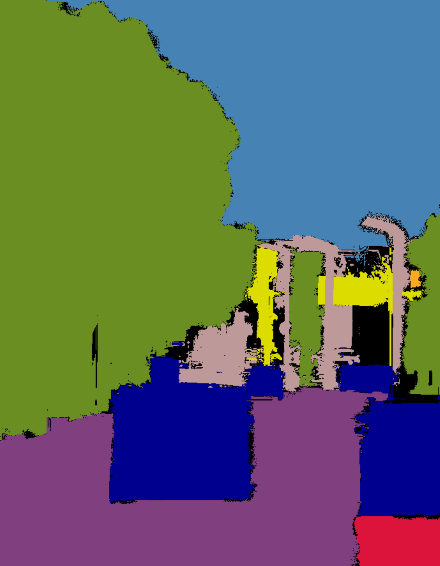}
				\end{minipage}
			}
			\hfill
			\subfloat[\centering Ours(refined label)]{
				\begin{minipage}[b]{0.12\linewidth}
					\centering
					\includegraphics[width=\linewidth]{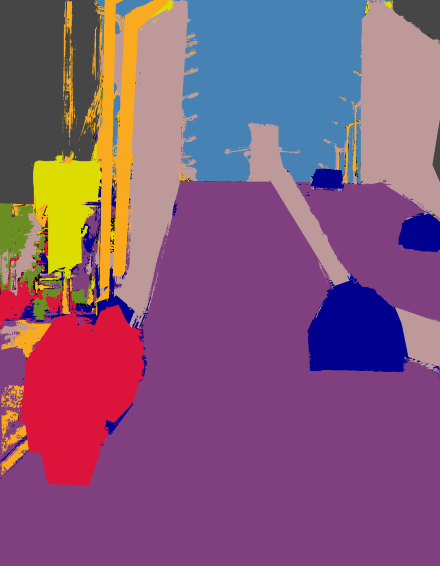}\vspace{4pt}
					\includegraphics[width=\linewidth]{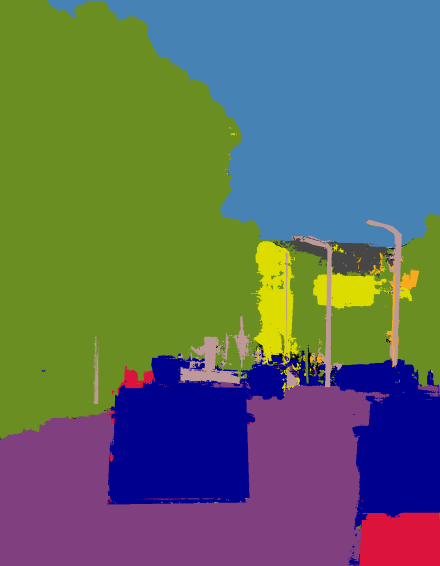}
				\end{minipage}
			}
		\end{minipage}
		\vfill
		\includegraphics[width=\linewidth]{images/class-eps-converted-to.pdf}
		
		\caption{Visualization results on Hyperspectral City V1.0 validation set. First we show the results of spectral semantic prior module based on RGB image and HSI respectively. Then we compare the results of semantic fusion module with coarse-to-fine annotation enrichment method.}
		
		\label{fig:refined label}
	\end{figure*}
	
	\begin{table*}[t]
		\caption{Comparisons of the results of hyperspectral semantic prior module and semantic fusion module with coarse-to-fine annotation enrichment \cite{luo2018coarse-to-fine} on Hyperspectral City V1.0 validation set $w.r.t$ mIoU, mean Acc. and IoU of each class.}
		\vspace{-6pt}
		\centering
		\begin{small}
			\begin{tabular}{c|c|c|c c c c c c c c c}
				\hline
				& mIoU & Acc. & car & human & road & light & sign & tree & building & sky & object\\
				\hline
				Coarse label       & 66.3 & 77.3 & 80.17 & \textbf{58.47} & 86.52 & \textbf{26.05} & 67.98 & 80.41 & 80.53 & 73.11 & 43.77 \\
				\hline
				Coarse-to-fine(0.3) & 63.6 & 80.8 & 80.17 & 46.91 & 87.72 & 23.45 & 62.34 & 82.05 & 81.27 & 75.03 & 33.56 \\
				Coarse-to-fine(0.7) & 64.4 & 80.3 & 80.85 & 48.60 & 87.57 & 24.13 & 66.27 & 81.96 & 81.43 & 74.66 & 34.16 \\
				Coarse-to-fine + HSI & 65.0 & 80.5 & 80.64 & 49.94 & 87.43 & 23.94 & 64.58 & 82.23 & 81.92 & 74.67 & 39.59 \\
				\hline
				Spectral prior (RGB) & 28.2 & 43.3 & 18.80 & 2.31 & 40.45 & 4.05 & 18.03 & 62.98 & 19.47 & 75.51 & 11.93 \\
				Spectral prior (HSI) & 54.2 & 77.0 & 47.93 & 46.62 & 81.07 & 14.74 & 47.99 & 72.83 & 58.48 & \textbf{90.96} & 27.33 \\
				\tabincell{c}{Refined label}      & \textbf{69.4} & \textbf{82.1} & \textbf{81.90} & 56.50 & \textbf{88.33} & 24.83 & \textbf{70.97} & \textbf{83.81} & \textbf{82.63} & 88.75 & \textbf{46.85} \\
				\hline
			\end{tabular}
		\end{small}
		\vspace{-6pt}
		\label{table:Refined label on validation set}
	\end{table*}
	%-------------------------------------------------------------------------
	\subsection{Implementation detail}
	\noindent\textbf{Hyperspectral City dataset.} %The Hyperspectral City V1.0 dataset is a novel spectral dataset for urban scene understanding. There are 10 classes are used for parsing evaluation. This dataset contains 58 high quality finely annotated images and 367 coarsely annotated images. Coarsely annotated images and the finely annotated images are used for training and testing. There are 6 images have both fine annotation and coarse annotation, which is used for validation. \newline
	The Hyperspectral City V1.0 dataset has 367 frames with coarse labels and 55 frames with fine labels. Coarsely and finely annotated images are used for training and testing respectively. There are 6 images have both fine and coarse annotations, which are used for validating. %Due to storage limitations, the number of images is limited but the scenes are rich enough. %The environment includes crowded traffic area, famous buildings and structures, central business district, highways, quiet suburbs and overpasses. The lighting conditions include day, night and sunset. 
	Spectral camera (PMVIS \cite{cao2011high}) can capture RGB and spectral images of the same spatial region at the same time. Therefore, each frame captures both the RGB image and the HSI, which have the same spatial resolution 1379 by 1773 and same label. The HSI has 129 spectral channels. The spectrum range is 450 to 950 nm (visible and near-infrared bands) and spectral resolution is 4nm.
	
	\noindent\textbf{Spectral prior.} 
	ResNet50 is adopted as the hyperspectral image classification network. Verified by experiment, we set the initial learning rate as 0.01, weight decay as 0.0005 and epoch as 30.\par
	
	Because HSIs consume a lot of memory, we prepare the data in two steps to balance memory and network training. First, we choose images from training dataset with the batch size 6. Second, we randomly select 10,000 pixels from one HSI, excluding whose corresponding coarse label is "0"(background). Each pixel generates a HSI cube, totally 60,000. Then we randomly choice cubes from these HSI cubes for training with the batch size 256. This method allows us to maximize the use of memory and prevents overfitting at one HSI. The number of HSI cubes used in each image is a small fraction of the total. The spectral information of the same substance has high similarity, which can ensure that sufficient prior information can be learned by using few HSI cubes. We set the spatial resolution of the HSI cube is $11\times11$.\newline
	\noindent\textbf{Refined label.}
	After generating spectral prior. We use the method at section 4.3 to generate fusion label. We generate the mask by eroding each class of coarse label. Ablation experiments are performed on selecting the optimal erosion kernel size and noise control threshold. Then we generate the refined labels on training dataset for finetuning module.\newline
	\textbf{Finetune network.}
	After getting refined label, we use refined label to fine-tune segmentation pre-trained model. For fine-tuning network, we fix the parameters of feature extraction layers, and only fine-tune the last two $1\times 1$ convolution layers. We set the initial learning rate as 0.001, weight decay as 0.0005, crop size as $1773\times1379$, epoch as 200 and batch size as 3 on four GPUs (GTX 1080Ti). We perform the polynomial learning rate policy with factor $1-(\frac{iter}{iter_{max}})^{0.9}$. We use $InPlace-ABN^{sync}$ ~\cite{bulo2018in-place} to synchronize the mean and standard-deviation of BN across multiple GPUs. For the data augmentation, we perform random flipping horizontally and random brightness. For evaluation, we use class-wise intersection over union $(IoU)$ and pixel-wise accuracy $(Acc.)$ metrics.
	
	%\begin{equation}
	%IoU = \frac{TP}{TP+FP+FN} \ , \label{5}
	%\end{equation}
	%where $TP$, $FP$, and $FN$ are the number of true positive, false positive, and false negative pixels, respectively, determined over the whole test set. %In the process of fine-tuning, we test the model on the validation set at the end of each epoch. After 200 epoch, we chose the model with the highest IoU. \par
	%-------------------------------------------------------------------------
	\subsection{Quantitative Results}
	
	%In this section, we mainly introduce the experiment of refined label and fusion label generation. Through this experiment, we get the label with the best fineness. \par

	%Here, we use HSI classification method to generate refined labels based on HSI and RGB image respectively. Then we compare the generated label and coarse label with the ground-truth dense annotation at validation set at the pixel-level. IoU and Acc. are applied to measure the percentage of correctly labelled pixels. 
	%This ablation experiment illustrates that in the acquisition stage of RGB image, a lot of spectral information has been lost.  \par
	\begin{figure*}[ht] 
		\centering 
		\begin{minipage}[b]{0.9\linewidth} 
			\centering
			\subfloat[Raw RGB]{
				\begin{minipage}[b]{0.16\linewidth} 
					\centering
					\includegraphics[height=1in]{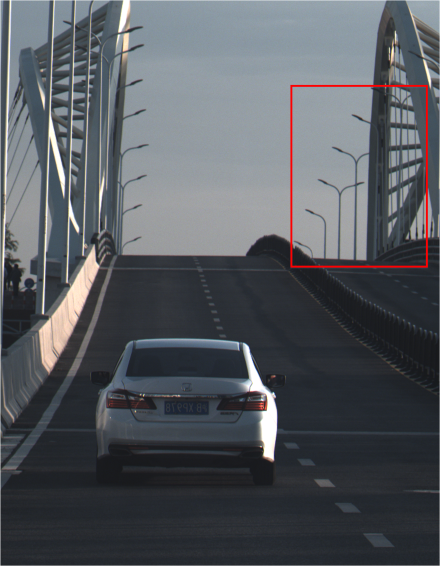}\vspace{4pt}
					\includegraphics[height=1in]{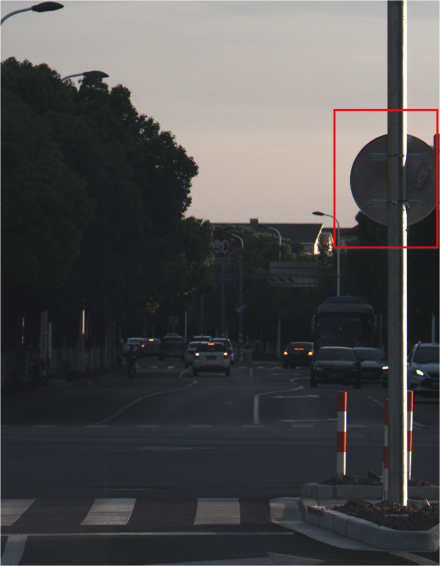}\vspace{4pt}
					\includegraphics[height=1in]{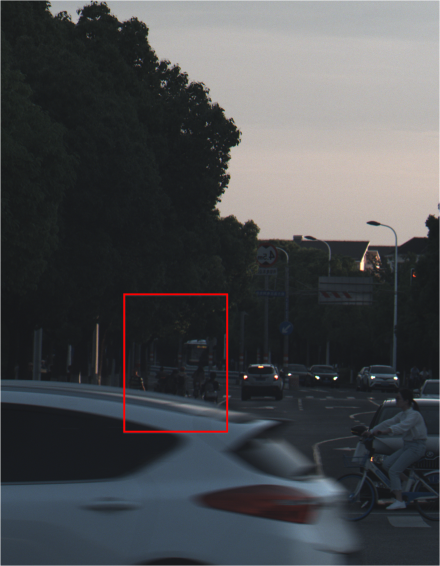}
				\end{minipage}
			}
			\subfloat[Coarse label]{
				\begin{minipage}[b]{0.16\linewidth}
					\centering
					\includegraphics[height=1in]{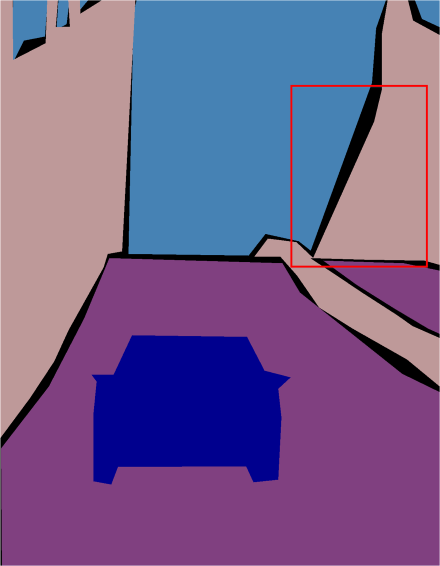}\vspace{4pt}
					\includegraphics[height=1in]{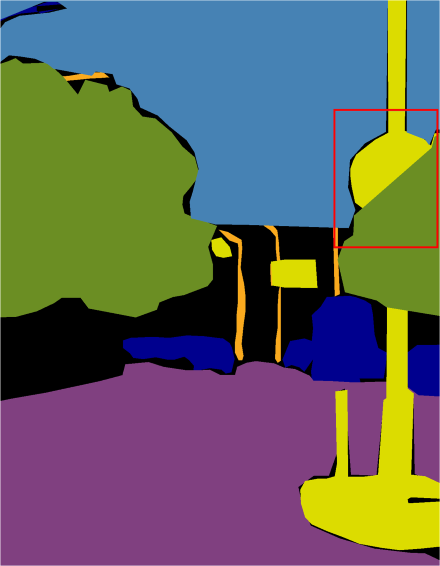}\vspace{4pt}
					\includegraphics[height=1in]{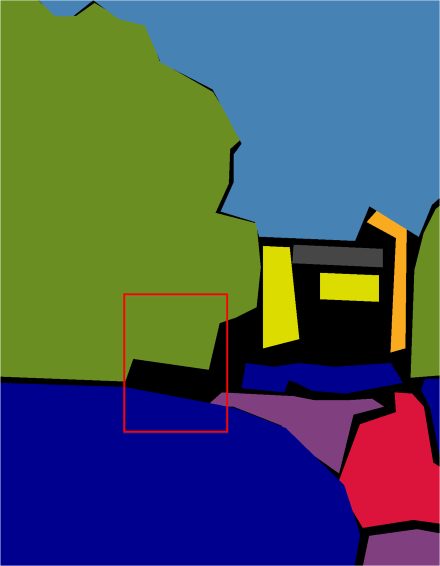}
				\end{minipage}
			}
			\subfloat[Spectral prior]{
				\begin{minipage}[b]{0.16\linewidth}
					\centering
					\includegraphics[height=1in]{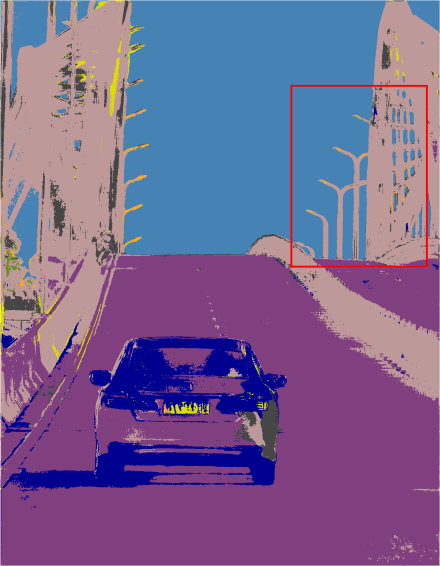}\vspace{4pt}
					\includegraphics[height=1in]{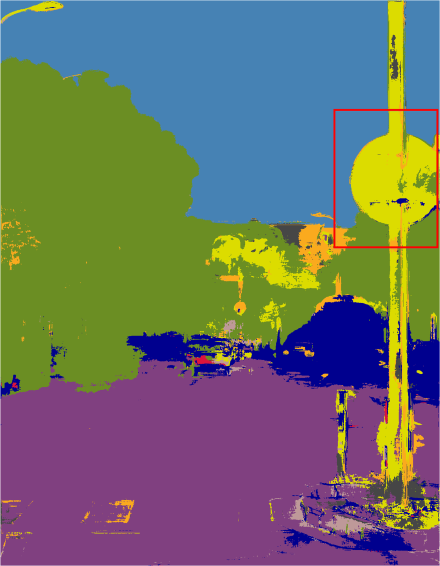}\vspace{4pt}
					\includegraphics[height=1in]{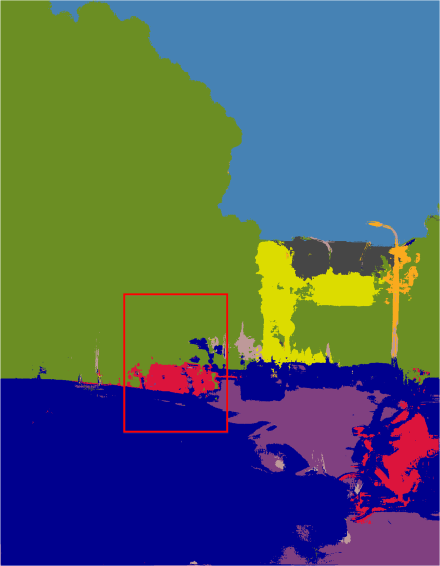}
				\end{minipage}
			}
			\subfloat[Raw RGB]{
				\begin{minipage}[b]{0.16\linewidth} 
					\centering
					\includegraphics[height=1in]{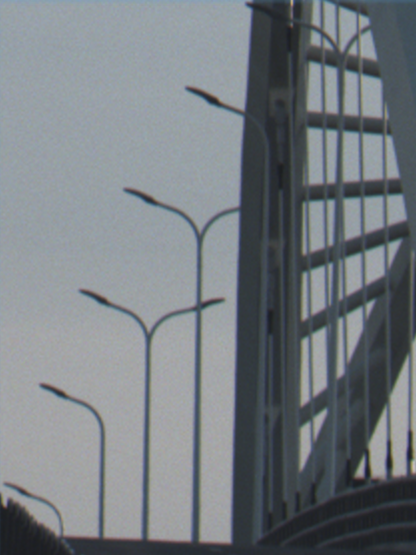}\vspace{4pt}
					\includegraphics[height=1in]{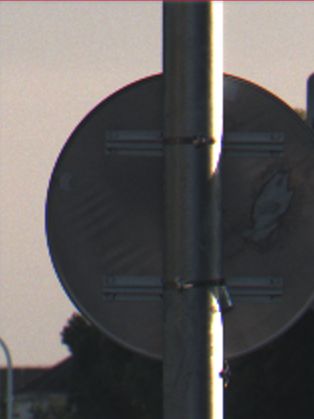}\vspace{4pt}
					\includegraphics[height=1in]{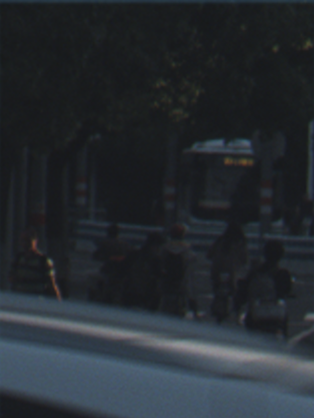}
				\end{minipage}
			}
			\subfloat[Coarse label]{
				\begin{minipage}[b]{0.16\linewidth}
					\centering
					\includegraphics[height=1in]{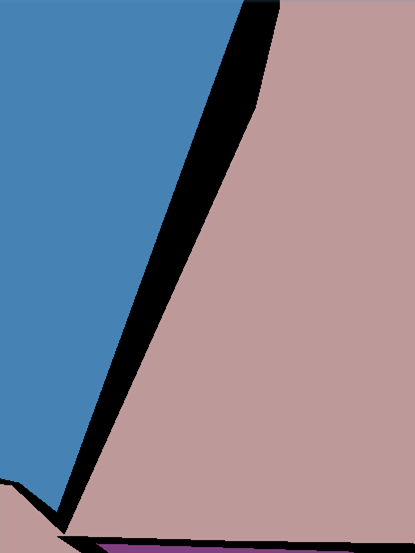}\vspace{4pt}
					\includegraphics[height=1in]{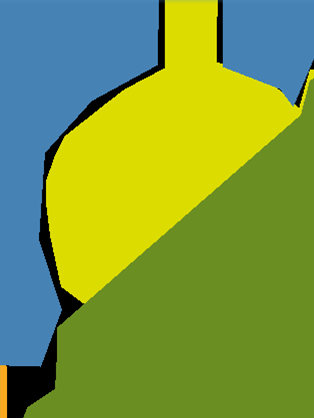}\vspace{4pt}
					\includegraphics[height=1in]{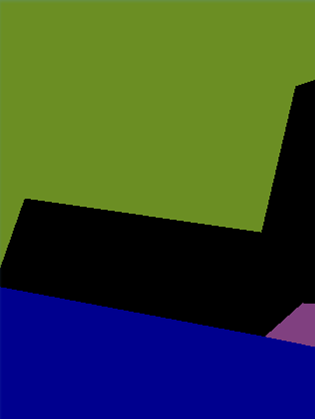}
				\end{minipage}
			}
			\subfloat[Spectral prior]{
				\begin{minipage}[b]{0.16\linewidth}
					\centering
					\includegraphics[height=1in]{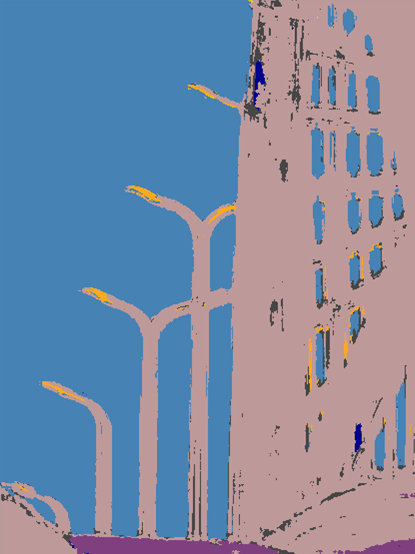}\vspace{4pt}
					\includegraphics[height=1in]{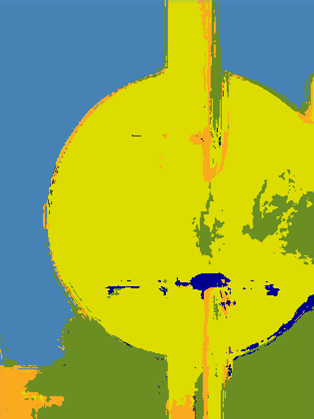}\vspace{4pt}
					\includegraphics[height=1in]{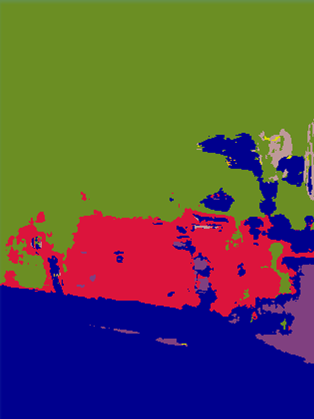}
				\end{minipage}
			}
		\end{minipage}
		\vfill
		\vspace{-6pt}
		\caption{Comparisons of spectral prior based on HSI on Hyperspectral City training set. The first 3 columns are the full image and label, and the last 3 columns are area zooms. Spectral prior improves accuracy and corrects errors compared to coarse labels.}
		\vspace{-12pt}
		\label{fig:refined label training}
	\end{figure*}
	
	\begin{figure}[htbp] 
		\centering 
		\begin{minipage}[b]{\linewidth} 
			\subfloat[RGB image]{
				\begin{minipage}[b]{0.17\linewidth} 
					\centering
					\includegraphics[width=\linewidth]{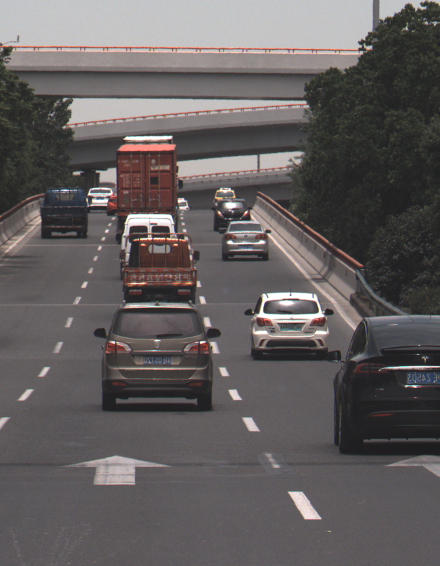}\vspace{4pt}
					\includegraphics[width=\linewidth]{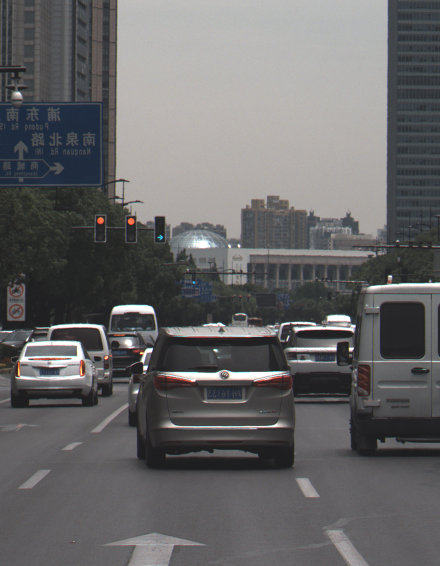}\vspace{4pt}
					\includegraphics[width=\linewidth]{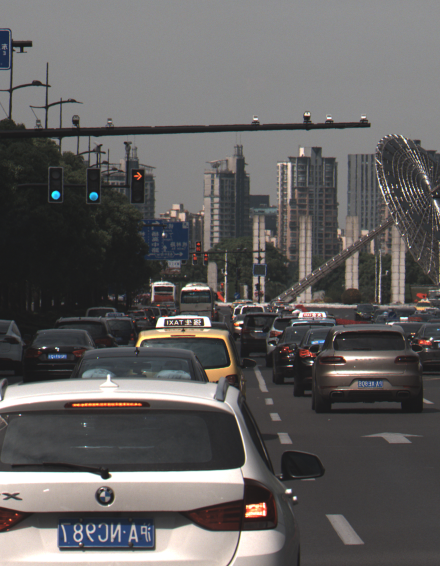}
				\end{minipage}
			}
			\hfill
			\subfloat[Ground-truth]{
				\begin{minipage}[b]{0.17\linewidth}
					\centering
					\includegraphics[width=\linewidth]{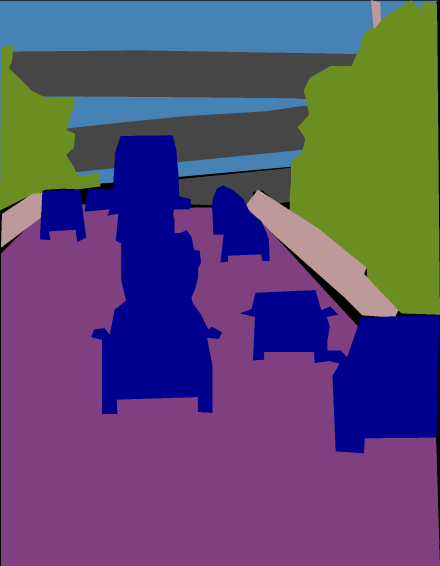}\vspace{4pt}
					\includegraphics[width=\linewidth]{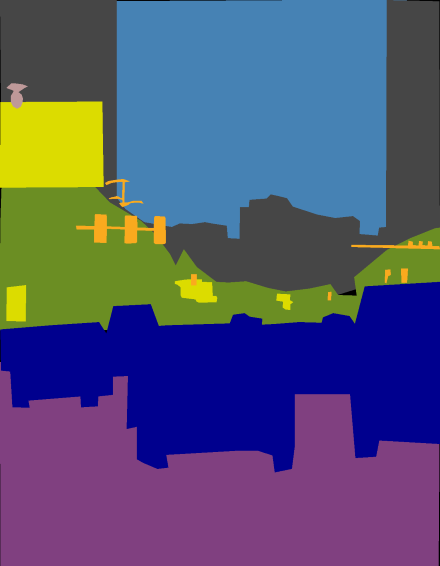}\vspace{4pt}
					\includegraphics[width=\linewidth]{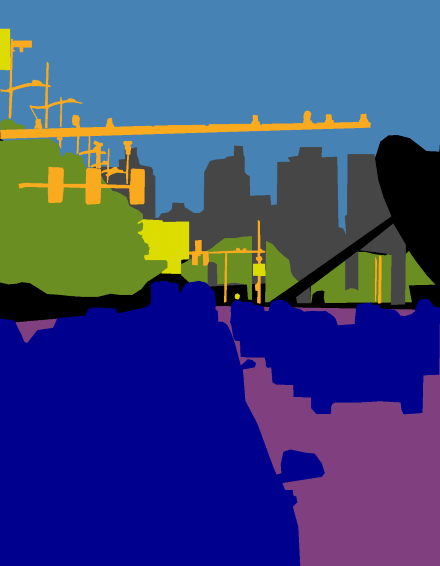}
				\end{minipage}
			}
			\hfill
			\subfloat[\centering Pre-trained model]{
				\begin{minipage}[b]{0.17\linewidth}
					\centering
					\includegraphics[width=\linewidth]{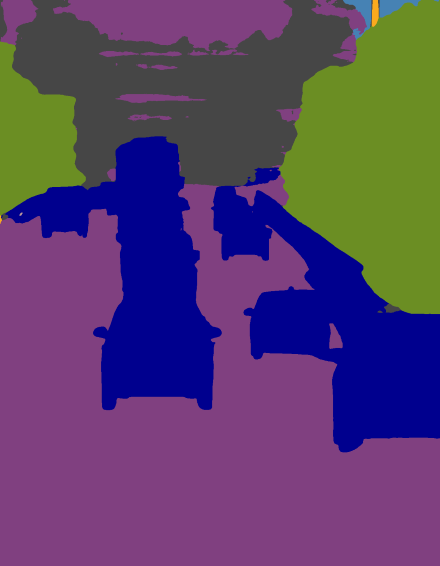}\vspace{4pt}
					\includegraphics[width=\linewidth]{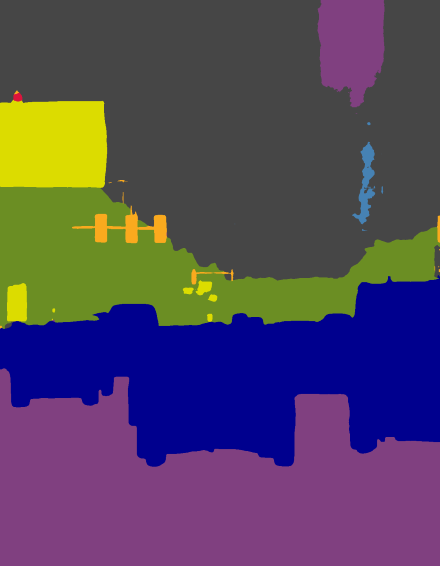}\vspace{4pt}
					\includegraphics[width=\linewidth]{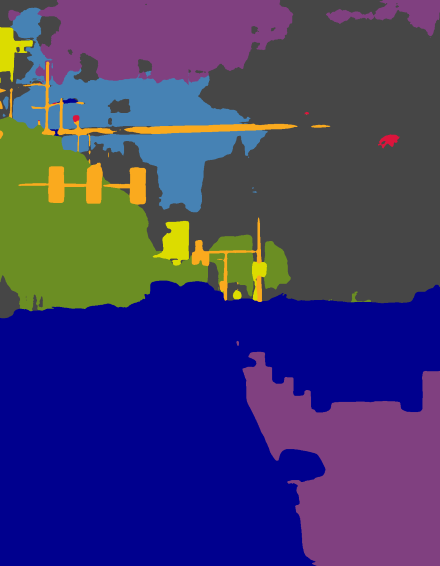}
				\end{minipage}
			}
			\hfill
			\subfloat[\centering Coarse label]{
				\begin{minipage}[b]{0.17\linewidth}
					\centering
					\includegraphics[width=\linewidth]{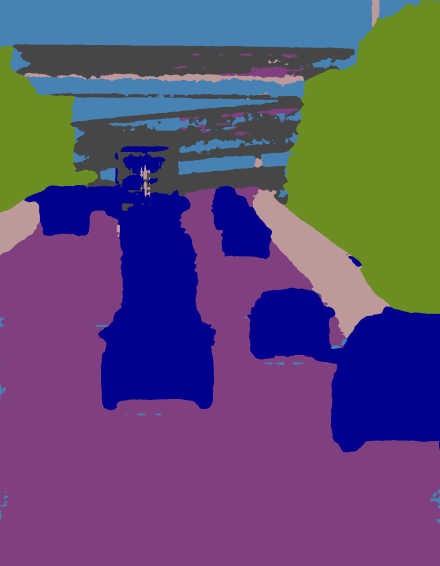}\vspace{4pt}
					\includegraphics[width=\linewidth]{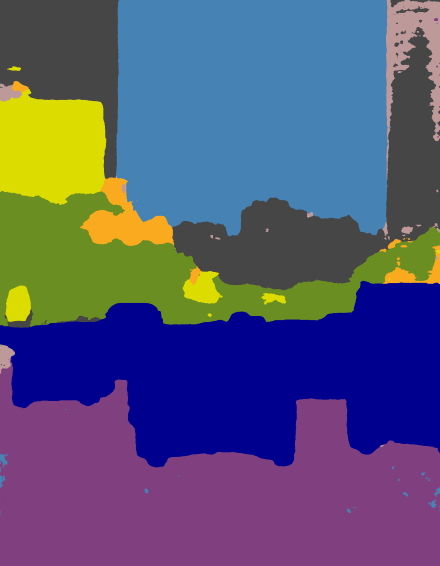}\vspace{4pt}
					\includegraphics[width=\linewidth]{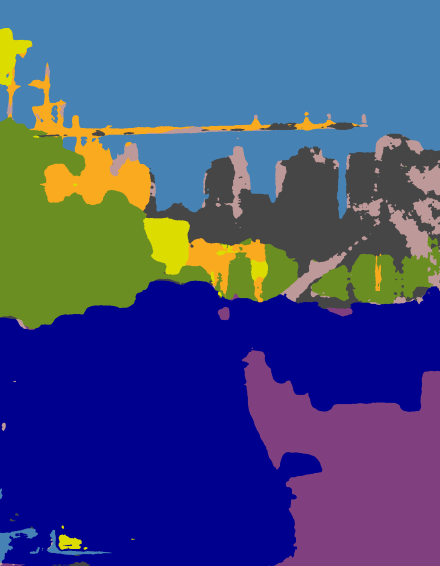}
				\end{minipage}
			}
			%\hfill
			%\subfloat[\centering Fine-tune with refined label]{
			%	\begin{minipage}[b]{0.15\linewidth}
			%		\centering
			%		\includegraphics[width=\linewidth]{images/r2.png}\vspace{4pt}
			%		\includegraphics[width=\linewidth]{images/r4.png}\vspace{4pt}
			%		\includegraphics[width=\linewidth]{images/r6.png}
			%	\end{minipage}
			%}
			\hfill
			\subfloat[\centering Refined label]{
				\begin{minipage}[b]{0.17\linewidth}
					\centering
					\includegraphics[width=\linewidth]{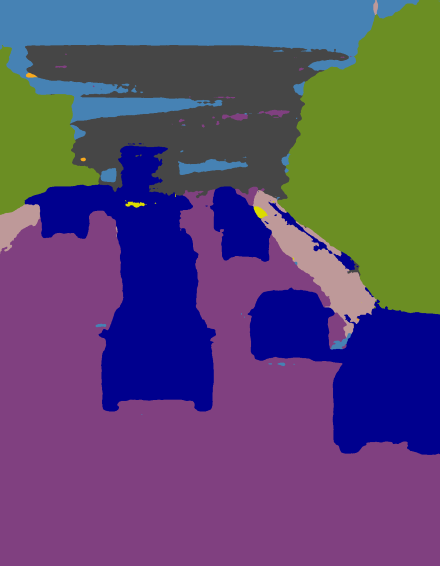}\vspace{4pt}
					\includegraphics[width=\linewidth]{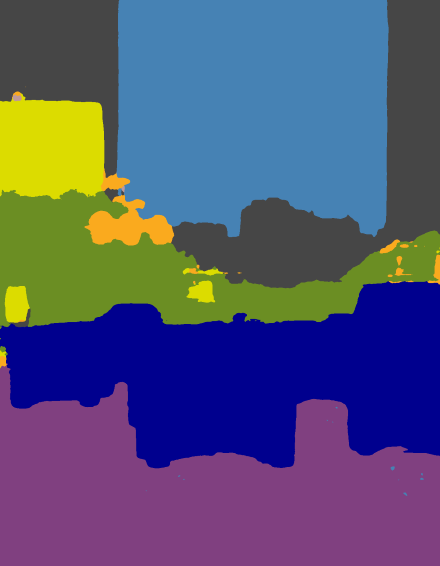}\vspace{4pt}
					\includegraphics[width=\linewidth]{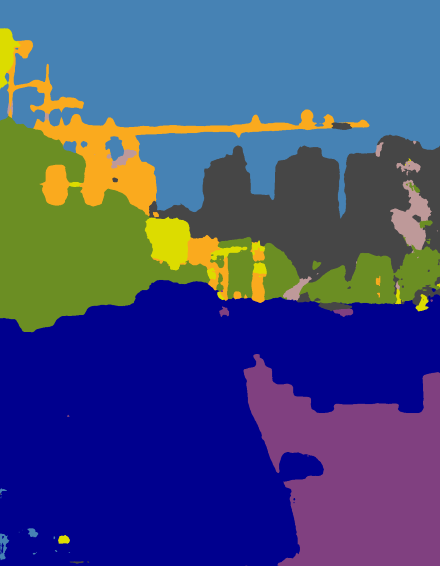}
				\end{minipage}
			}
		\end{minipage}
		\vfill
		\includegraphics[width=\linewidth]{images/class-eps-converted-to.pdf}
		\vspace{-6pt}
		\caption{Semantic segmentation results of fine-tuning HRNet pre-trained model under different supervision on Hyperspectral City testing set.}
		\vspace{-18pt}
		\label{fig:refine label training}
	\end{figure}
	
	\noindent\textbf{Spectral prior.}
	In Table \ref{table:Refined label on validation set}, we compare the spectral prior with coarse label at validation set. 
	%We compute IoU and Acc. for all the pixel and Acc. of each class. 
	First, the spectral prior based on the HSIs is much better than that based on the RGB images. HSIs have stronger semantic prior than RGB images. 
	%Good segmentation accuracy can be obtained by using only the spectral information and a little space information of hyperspectral images. This proves that the spectral information has significant  value in the automatic driving scene. 
	Second, because the spectral prior has misclassification within some classes, spectral prior is almost equal to coarse labels in Acc. and lower in mIoU. 
	Third, Figure \ref{fig:refined label} and Figure \ref{fig:refined label training} are the comparisons of some spectral prior and coarse labels on the validation set and training set, which show that spectral prior can effectively improve the edge fineness and correct the wrong or missing annotation in the coarse labels.
	%For example, row 1, 5, the refined label significantly improved the labeling accuracy compared with the coarse label. Row 2, 3, 4, for some unlabeled classes or mislabeled areas, refined labels can also be a good supplement and correction. The refined label based on HSI classification module, on the one hand, can effectively improve the edge fineness of all kinds of labels, on the other hand, can correct the wrong labels and missing labels in the coarse labels. However, in the absence of spatial information, there will inevitably be some classification errors for some categories containing more complex internal substances. So in the next section we combine the advantages of the two labels to create the fusion label.\par
	
	%-------------------------------------------------------------------------
	\noindent\textbf{Refined label.} %In this work, to take advantage of coarse labels and refined labels, we present the proposed fusion label method. 
	We compare the best refined labels with the original coarse labels, spectral prior and the result of coarse-to-fine annotation enrichment method \cite{luo2018coarse-to-fine}. IoU and Acc. are applied to measure the percentage of correctly labeled pixels. Noise control and class-wise erosion fusion are applied to generate refined labels. The parameters selection will be described in the ablation study. \par
	As Table \ref{table:Refined label on validation set} shows, the refined label has the highest Acc.($82.1\%$) and mIoU ($69.4\%$) which brings $4.8\%$ and $3.1\%$ improvement than coarse label. Compare with coarse-to-fine annotation enrichment method \cite{luo2018coarse-to-fine}, our refined label achieves $5.0\%$ mIoU and $1.8\%$ Acc. absolute gains. As Figure \ref{fig:refined label} shows, due to the coarse-to-fine method relies heavily on the coarse labeled region and does not fit well on large background region, the mIoU score is even lower than coarse label. To further compare with coarse-to-fine method, we add spectral information to the coarse-to-fine method. As Table \ref{table:Refined label on validation set} shows, HSIs can bring $0.6\%$ mIoU improvement. But the results are still weaker than ours. Compared with existing methods, our method has lower requirements on the quality of coarse labels and achieves better results.
	%Manual annotation can ensure the correctness of most annotation areas, but it lacks enough fine granularity in the category boundary area and is easy to produce errors. For semantic segmentation based on RGB image, the fineness of tag edge is very important. Because RGB image has less spectral information and more spatial structure information, the prior relationship between spatial structure information and category depends on detailed annotation. Therefore, in this case, our proposed label refinement method can make up for this shortcoming. The fusion label based on spectral information can has a better segmentation precision at the classification edge and correct some mislabeling. \par
	%-------------------------------------------------------------------------
	
	\begin{table}[t]
		\caption{Results of fine-tuning HRNet pre-trained on Cityscapes with different supervision for semantic segmentation on Hyperspectral City testing set.}
		\vspace{-6pt}
		\centering
		\begin{small}
			\begin{tabular}{c|c|c|c}
				\hline
				&Supervision &mIoU(\%) & Acc(\%) \\
				\hline
				\multirow{3}{*}{HRNet}&Pre-trained model & 59.30 & 78.57 \\
				&Coarse label                           & 67.24 & 87.21 \\
				&Refined label                           & \textbf{69.40} & \textbf{89.12} \\
				\hline
				\multirow{3}{*}{DeepLabV3+}&Pre-trained model & 55.66 & 79.04 \\
				&Coarse label                           & 61.90 & 85.23 \\
				&Refined label                           & \textbf{63.36} & \textbf{85.71} \\
				\hline
			\end{tabular}
		\end{small}
		\vspace{-6pt}
		\label{table:finetune result}
	\end{table}
	
	\begin{table*}[t]
		\caption{Comparisons of refined label with erosion kernel size $l$ on Hyperspectral City validation set $w.r.t$ mIoU. The threshold of noise control is 0.7. For each class, we select the kernel size $l_i$ with the highest mIoU from 1 to $n$.}
		\vspace{-6pt}
		\centering
		\begin{small}
			\begin{tabular}{c|c|c c c c c c c c c}
				\hline
				kernel size $l$ & mIoU(\%)  & car & human & road & light & sign & tree & building & sky & object\\
				\hline
				Baseline(n=1)    & 69.20  & 1 & 1 & 1 & 1 & 1 & 1 & 1 & 1 & 1 \\
				n=5  & 69.33  & 1 & 1 & 5 & 1 & 5 & 5 & 3 & 5 & 5 \\
				n=9  & 69.37  & 1 & 1 & 7 & 1 & 9 & 5 & 3 & 9 & 9 \\
				n=11 & \textbf{69.41}  & 1 & 1 & 7 & 1 & 11 & 5 & 3 & 11 & 11 \\
				n=15 & 69.39  & 1 & 1 & 7 & 1 & 13 & 5 & 3 & 13 & 13 \\
				\hline
			\end{tabular}
		\end{small}
		\vspace{-6pt}
		\label{table:erosion kernel size}
	\end{table*}
	
	%In this section, to demonstrate the value of fusion labels in actual semantic segmentation, we use fusion labels to finetune HRNet pertrained model.
	\noindent\textbf{Finetune network analysis.}
	As shown in Table \ref{table:finetune result},
	%Cityscapes is a dataset of urban road scenes widely used in semantic segmentation, which has good mobility. HRNet is a strong backbone for computer vision problems and has superior result near the SOTA result on Cityscapes. 
	first we use HRNet (HRNetV2-W48) pertrained on Cityscapes as the baseline, whose mIoU is 81.1\% on Cityscapes. We divided 19 categories of Cityscapes into 10 categories in the Hyperspectral City by class affiliation. HRNet per-trained model achieves 59.30\% mIoU. Although the pre-trained model based on the Cityscapes dataset with fine annotation has high segmentation accuracy, the results of the direct migration pretrained model are poor. We use coarse label to train the network. Coarse label gives network some semantic supervision, but it will weaken the precision of pre-trained model. As shown in Table \ref{table:finetune result}, we use refined label to fine-tune HRNet and achieve the best mIoU (69.40\%) and Acc. (89.12\%). We also use DeepLabV3+ (MobileNetV2 as backbone) for fine-tuning module in Table \ref{table:finetune result}. Although our fine-tuning fixed most of the parameters of network, the results also show that our method can bring great segmentation performance improvement.\par
	%Experimental results show that our method can effectively improve the quality and classification accuracy of labels with the help of spectral information.
	
	%-------------------------------------------------------------------------
	\subsection{Further Ablation Study}
	
	\begin{table}[t]
		\caption{Comparison of spectral prior generated from different HSI cube sizes on Hyperspectral City validation set.}
		\vspace{-6pt}
		\centering
		\begin{small}
			\begin{tabular}{c|c|c}
				\hline
				HSI cube size &mIoU(\%) & Acc(\%) \\
				\hline
				$5\times 5$      & 42.04 & 75.43 \\
				$11\times 11$    & \textbf{54.21} & \textbf{76.95} \\
				$25\times 25$    & 42.30 & 74.95 \\
				
				\hline
			\end{tabular}
		\end{small}
		\label{table:HSI cube size result}
		\vspace{-0pt}
	\end{table}
	
	%To analyze the effectiveness of our coarse label refinement framework, we conduct ablation study with different designs. In this ablation study we aim to exam the results under different experimental parameters and to ensure that we achieve the best experimental results on this dataset. \par
	\noindent\textbf{Spectral prior.} %In order to avoid the influence of coarse labeling spatial information, hyperspectral labels are generated by processing hyperspectral images into hyperspectral cubes of a certain size. Hyperspectral cubes introduce spatial information and spectral information within the neighborhood of the center pint. Therefore, 
	Smaller hyperspectral cubes contain too little spectral information, while larger hyperspectral cubes contain too much coarse spatial information, which all will affect the classification accuracy. Two kinds of information should be taken into account in the selection of hyperspectral cube size. As shown in Table \ref{table:HSI cube size result},
	%In order to compare the influence of different sizes of HSI cube on the generation of refined labels,
	three spectral cube sizes (5, 11 and 25) are compared. The HSI cubes of size 11 achieve the highest mIoU and Acc. on the validation set.%, which indicated that this size contained sufficient spectral information while avoiding the interference of coarse labels
	
	\noindent\textbf{Noise control and Class-wise erosion.} In this ablation study, we give comparisons of the noise control and class-wise erosion fusion. Since noise control will introduce few unlabeled area, mIoU is more suitable as an evaluation index of label quality.\par
\begin{figure}[t]
		\centering
		\includegraphics[width=1\linewidth]{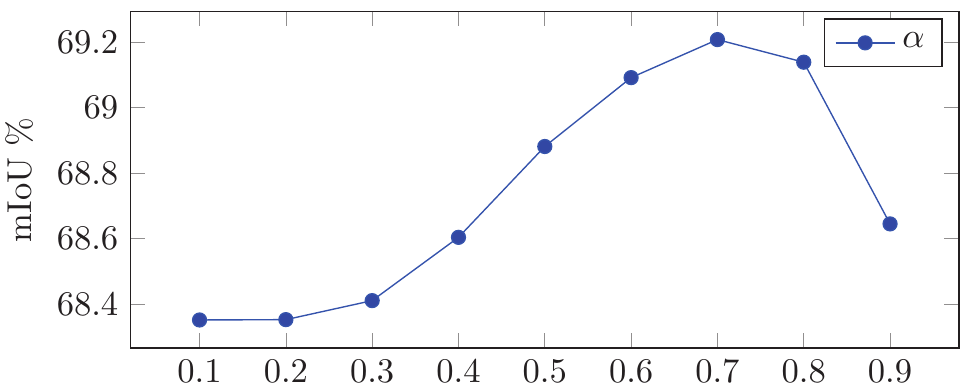}
		\vspace{-12pt}
		\caption{Ablation study of threshold $\alpha$ in noise control. }
		\label{fig:threshold}
		\vspace{-6pt}
	\end{figure}

	First, we test the threshold $\alpha$ of noise control. The spectral prior is directly fused with the coarse label after the noise control operation. After generating the refined label, the mIoU scores under different thresholds are tested on the ground truth of the validation set. As shown in Figure \ref{fig:threshold}, we test $\alpha$ from 0.1 to 0.9, and find that refined label with $\alpha=0.7$ achieves the best result. 
	Then, we test the erosion kernel size $l$ of class-wise erosion. Firstly The spectral prior for fusion is subjected to noise control with $\alpha = 0.7$. Then we use the class-wise erosion fusion mentioned in section 4.3 to fuse spectral prior and coarse label. As shown in Table \ref{table:erosion kernel size}, the refined label achieves highest mIoU $(69.41\%)$ with n=11. Class-wise erosion selects the most appropriate annotation area for different class, which better combines the advantages of the two kinds of labels. The results show that class-wise erosion fusion can further improve the performance of refined label.
	Finally, we give the qualitative comparisons on Table \ref{table:CE and NS}, which demonstrates that operations in semantic fusion module can improve the performance of the refined label respectively. 
	
	\begin{table}[t]
		\caption{Comparisons of semantic fusion module methods (CE: class-wise erosion(n=11), NS: noise control($\alpha$ = 0.7)) with fine label of Hyperspectral City validation set $w.r.t$ mIoU.}
		\vspace{-6pt}
		\centering
		\begin{small}
			\begin{tabular}{c|c c|c }
				\hline
				{}& CE & NS & mIoU(\%)  \\
				\hline
				Coarse label& & & 66.3  \\
				\hline
				\multirow{4}{*}{Refined label}& \XSolidBrush & \XSolidBrush & 68.3  \\
				& \Checkmark   & \XSolidBrush & 68.4  \\
				& \XSolidBrush & \Checkmark & 69.2  \\
				& \Checkmark   & \Checkmark & \textbf{69.4} \\
				\hline
			\end{tabular}
		\end{small}
		\label{table:CE and NS}
		\vspace{-12pt}
	\end{table}
	
	%-------------------------------------------------------------------------
	\vspace{-6pt}
	\section{Conclusion}
	\vspace{-6pt}
	In this paper, we present a weakly-supervised semantic segmentation framework via HSI based on hyperspectral cityscape scenes. Specifically, first, we introduce a new hyperspectral dataset. The comparisons between hyperspectral images (HSIs) and RGB images prove that richer spectral information of HSIs is important to semantic prior. Second, we use the character that spectral information is independent of fine annotation to optimize the semantic segmentation coarse annotation. The label with higher precision is obtained in the case of lower annotation cost. Third, we use the refined label to finetune the semantic segmentation pre-trained model, which significantly improves the segmentation accuracy.
	%Experimental results show that our method is simple enough and has good mobility, which can adjust the generation of mask for different data set labels to achieve the optimal refinement label. The annotation cost of data sets can be greatly reduced. At the same time, this method proves the important value of hyperspectral information for automatic driving system and semantic segmentation. 
	In future, we hope to continue to explore the application value of spectra in more scenes.
	%On the other hand, since hyperspectral images contain information of near-infrared bands, we hope to explore the application value of different bands to semantic segmentation.
	%-------------------------------------------------------------------------
	
	{\small
		\bibliographystyle{ieee_fullname}
		\bibliography{egbib}
	}
	
\end{document}